%% file: main.tex
\documentclass[10pt,journal]{IEEEtran}
\IEEEoverridecommandlockouts
\usepackage{amsmath,amsfonts}
\usepackage[noend]{algpseudocode}
\usepackage{algorithm}
\usepackage{array}
\usepackage[font=footnotesize]{caption} 
\usepackage{textcomp}
\usepackage{stfloats}
\usepackage{url}
\usepackage{verbatim}
\usepackage{subfigure}
\usepackage{graphicx,subcaption}
\usepackage{cite}
\usepackage{threeparttable}

\usepackage{xcolor}
\usepackage{multirow,booktabs}
\usepackage{makecell}
\usepackage{enumitem}
\usepackage{bm}
\usepackage{amssymb}
\usepackage{amsthm}

\usepackage{soul}

\theoremstyle{plain}
\newtheorem{thm}{Theorem}

\newtheorem{prop}[thm]{Proposition}

\allowdisplaybreaks[4]

\begin{document}

\bstctlcite{IEEEexample:BSTcontrol}

\title{Implicit Neural Compression of Point Clouds}

\author{Hongning Ruan, Yulin Shao, Qianqian Yang, Liang Zhao, Zhaoyang Zhang, Dusit Niyato
\thanks{H. Ruan, Q. Yang, L. Zhao and Z. Zhang are with the College of Information Science and Electronic Engineering, Zhejiang University (e-mails: \{rhohenning, qianqianyang20, lzhao2020, ning\_ming\}@zju.edu.cn).

Y. Shao is with the Department of Electrical and Electronic Engineering, The University of Hong Kong, Hong Kong, China (e-mail: ylshao@hku.hk).

D. Niyato is with the College of Computing and Data Science, Nanyang Technological University, Singapore (e-mail: dniyato@ntu.edu.cn).

{ The source code will be released at https://github.com/RhoHenning/NeRC3.}

An earlier version of this paper\cite{ruan2024point} was presented in part at the IEEE/CIC International Conference on Communications in China (ICCC), Hangzhou, China, in August 2024.
}}

\maketitle

\begin{abstract}
Point clouds have gained prominence across numerous applications due to their ability to accurately represent 3D objects and scenes. However, efficiently compressing unstructured, high-precision point cloud data remains a significant challenge. In this paper, we propose NeRC$^{\textbf{3}}$, a novel point cloud compression framework that leverages implicit neural representations (INRs) to encode both geometry and attributes { of dense point clouds. Our approach employs two coordinate-based neural networks: one maps spatial coordinates to voxel occupancy, while the other maps occupied voxels to their attributes, thereby implicitly representing the geometry and attributes of a voxelized point cloud. The encoder quantizes and compresses network parameters alongside auxiliary information required for reconstruction, while the decoder reconstructs the original point cloud by inputting voxel coordinates into the neural networks.} Furthermore, we extend our method to dynamic point cloud compression through techniques that reduce temporal redundancy, including a 4D spatio-temporal representation termed 4D-NeRC$^{\textbf{3}}$. Experimental results validate the effectiveness of our approach: { For static point clouds, NeRC$^{\textbf{3}}$ outperforms octree-based G-PCC standard and existing INR-based methods. For dynamic point clouds, 4D-NeRC$^{\textbf{3}}$ achieves superior geometry compression performance compared to the latest G-PCC and V-PCC standards, while matching state-of-the-art learning-based methods. It also demonstrates competitive performance in joint geometry and attribute compression.}
\end{abstract}


\section{Introduction}\label{sec:I}
\input{SecI}

\section{Related Works}\label{sec:II}
\input{SecII}

\section{INR-based Point Cloud Compression}\label{sec:III}
\input{SecIII}

\section{INR-based Dynamic Point Cloud Compression}\label{sec:IV}
\input{SecIV}

\section{Experimental Results}\label{sec:V}
\input{SecV}

\section{Conclusion}\label{sec:VI}
\input{SecVI}

\bibliographystyle{IEEEtran}
\bibliography{ref.bib}

\end{document}


\bstctlcite{IEEEexample:BSTcontrol}

\title{Implicit Neural Compression of Point Clouds: Supplementary Material}

\author{Hongning Ruan, Yulin Shao, Qianqian Yang, Liang Zhao, Zhaoyang Zhang, Dusit Niyato}

\maketitle

\begin{abstract}
This document provides additional implementation and experimental details.
\end{abstract}

\section{Additional Details of Dynamic PCC}

In this section, we present the additional details of dynamic PCC methods, i.e., r-NeRC$^3$ and c-NeRC$^3$.

\subsection{Notations}

Let $\mathcal{X}^{(t)}$ denote the geometry of the $t$-th frame, and $C^{(t)}$ denote the corresponding attributes that map each point $\mathbf{x}\in\mathcal{X}^{(t)}$ to its RGB color $C^{(t)}(\mathbf{x})$. In both r-NeRC$^3$ and c-NeRC$^3$, different frames are implicitly represented by separate neural networks. We use $\bm{\Theta}^{(t)}$ to denote the parameter set of network $F$ that represents the geometry of $t$-th frame $\mathcal{X}^{(t)}$, and use $\bm{\Phi}^{(t)}$ to denote the parameter set of network $G$ that represents the attributes $C^{(t)}$.

In Section III of the main paper (i.e., i-NeRC$^3$), we concentrate on the compression of static point clouds, where the variables and functions are defined on a single frame. Here, we extend these definitions and notations to multi-frame scenario by adding a superscript $(\cdot)^{(t)}$. For example, the geometry distortion loss is defined by the following expressions:
\begin{align}
D_F(\mathbf{x})&=-\widetilde{\alpha}(\mathbf{x})\cdot(1-\widetilde{p}(\mathbf{x}))^{\gamma}\log(\widetilde{p}(\mathbf{x})),\\
\widetilde{\alpha}(\mathbf{x})&=y(\mathbf{x})\cdot\alpha+(1-y(\mathbf{x}))\cdot(1-\alpha),\\
\widetilde{p}(\mathbf{x})&=y(\mathbf{x})\cdot p(\mathbf{x})+(1-y(\mathbf{x}))\cdot(1-p(\mathbf{x})),\\
y(\mathbf{x})&=\mathbb{I}(\mathbf{x}\in\mathcal{X})=\begin{cases}
1, & \mathbf{x}\in\mathcal{X}, \\
0, & \mathbf{x}\not\in\mathcal{X},
\end{cases}\\
p(\mathbf{x})&=F(\mathbf{x};\mathbf{\Theta}).
\end{align}

\noindent Ignoring the hyperparameters $\alpha,\gamma$ that remain constant across frames, we observe that $D_F(\mathbf{x})$ is defined on $\bm{\Theta}$ and $\mathcal{X}$. By replacing $\{\bm{\Theta},\mathcal{X}\}$ with $\{\bm{\Theta}^{(t)},\mathcal{X}^{(t)}\}$ and following the above relationship between variables and functions, we can derive the expression for the geometry distortion of the $t$-th frame, which we denote as $D_F^{(t)}(\mathbf{x})$.

As another example, the sample distribution for network $F$ is defined by the following expressions:
\begin{align}
\mathcal{P}_F&=\beta^*\cdot\mathcal{U}(\mathcal{X})+\alpha^*\cdot\mathcal{U}(\mathcal{V}),\\
\mathcal{V}&=\{\mathbf{x}:\lfloor\mathbf{x}/2^{N-M}\rfloor\in\mathcal{W},\mathbf{x}\in\mathcal{S}\},\\
\mathcal{W}&=\{\mathbf{w}:\mathbf{w}=\lfloor\mathbf{x}/2^{N-M}\rfloor,\mathbf{x}\in\mathcal{X}\}.
\end{align}

\noindent We observe that $\mathcal{P}_F$ is defined on the geometry $\mathcal{X}$. By replacing $\mathcal{X}$ with $\mathcal{X}^{(t)}$, we can sequentially derive $\mathcal{W}^{(t)}$, $\mathcal{V}^{(t)}$ and $\mathcal{P}_F^{(t)}$. 

\subsection{Additional Details of r-NeRC$^3$}

\subsubsection{Geometry Compression}
The first frame of a sequence is processed with the intra-frame compression method, as in i-NeRC$^3$. The encoder optimizes the parameters $\mathbf{\Theta}^{(0)}$, performs quantization, and then transmits the quantized parameters $\widehat{\mathbf{\Theta}}^{(0)}$ to the decoder. 

For the remaining frames ($t=1,2,\cdots$), the encoder first trains the complete parameters $\mathbf{\Theta}^{(t)}$ by minimizing the following loss function.
\begin{equation}\label{eq:rnerc3_loss_geom}
\mathcal{L}_F^{(t)}(\bm{\Theta}^{(t)})=\mathbb{E}_{\mathbf{x}\sim \mathcal{P}_F^{(t)}}[D_F^{(t)}(\mathbf{x})]+\frac{\lambda_F}{|\mathcal{X}^{(t)}|}\|\mathbf{\Theta}^{(t)}-\widehat{\mathbf{\Theta}}^{(t-1)}\|_1,
\end{equation}

\noindent where $\widehat{\mathbf{\Theta}}^{(t-1)}$ is the quantized parameters for the previous frame, which is frozen during the training process for the current frame. Then the encoder generates the residual by computing the difference between the parameters for the current frame and the quantized parameters for the previous frame:
\begin{equation}
\delta\mathbf{\Theta}^{(t)}=\mathbf{\Theta}^{(t)}-\widehat{\mathbf{\Theta}}^{(t-1)}.
\end{equation}

\noindent The encoder only quantizes and transmits the residual. The quantized residual can be expressed as $\delta\widehat{\mathbf{\Theta}}^{(t)}=\lfloor\delta\mathbf{\Theta}^{(t)}/\Delta_F\rceil\cdot\Delta_F$, where $\Delta_F$ is the step size.

The decoder retrieves the complete parameters by adding the residual to the buffered parameters for the previous frame, i.e.,
\begin{equation}
\widehat{\mathbf{\Theta}}^{(t)}=\widehat{\mathbf{\Theta}}^{(t-1)}+\delta\widehat{\mathbf{\Theta}}^{(t)}.
\end{equation}

These parameters are utilized to reconstruct the geometry for the current frame $\widehat{\mathcal{X}}^{(t)}$, as done in i-NeRC$^3$.

\subsubsection{Attribute Compression}

The first frame is processed as in i-NeRC$^3$, i.e., the encoder quantizes and transmits the complete parameters $\bm{\Phi}^{(0)}$. For the remaining frames, the encoder optimizes $\bm{\Phi}^{(t)}$ through the following loss function:
\begin{equation}\label{eq:rnerc3_loss_attr}
\mathcal{L}_G^{(t)}(\bm{\Phi}^{(t)})=\mathbb{E}_{\widehat{\mathbf{x}}\sim \mathcal{P}_G^{(t)}}[D_G^{(t)}(\widehat{\mathbf{x}})]+\frac{\lambda_G}{|\mathcal{X}^{(t)}|}\|\mathbf{\Phi}^{(t)}-\widehat{\mathbf{\Phi}}^{(t-1)}\|_1.
\end{equation}

Then the encoder quantizes and transmits the residual w.r.t. the previous frame, with details similar to geometry compression. The residual is computed as
\begin{equation}
\delta\mathbf{\Phi}^{(t)}=\mathbf{\Phi}^{(t)}-\widehat{\mathbf{\Phi}}^{(t-1)}.
\end{equation}

\noindent The quantized residual is $\delta\widehat{\mathbf{\Phi}}^{(t)}=\lfloor\delta\mathbf{\Phi}^{(t)}/\Delta_G\rceil\cdot\Delta_G$, where $\Delta_G$ is the step size.

The decoder retrieves the complete parameters by
\begin{equation}
\widehat{\mathbf{\Phi}}^{(t)}=\widehat{\mathbf{\Phi}}^{(t-1)}+\delta\widehat{\mathbf{\Phi}}^{(t)},
\end{equation}

\noindent which reconstructs the attributes $\widehat{C}^{(t)}$.

\subsection{Additional Details of c-NeRC$^3$}

\subsubsection{Geometry Compression}

Consider a group of consecutive $T$ frames $\mathcal{X}^{(0)},\mathcal{X}^{(1)},\cdots,\mathcal{X}^{(T-1)}$. Let $\bm{\Theta}^{(t)}$ be the parameter set of network $F$ that implicitly represents the $t$-th frame $\mathcal{X}^{(t)}$. The corresponding loss function that the parameter set $\mathbf{\Theta}^{(t)}$ minimizes is the geometry distortion averaged over training samples for the $t$-th frame, i.e., $\mathbb{E}_{\mathbf{x}\sim \mathcal{P}_F^{(t)}}[D_F^{(t)}(\mathbf{x})]$.

Let $\mathbb{R}^{|\mathbf{\Theta}|}$ denote the neural space, where $|\mathbf{\Theta}|$ is the number of parameters of network $F$. We assume that $\mathbf{\Theta}^{(0)},\mathbf{\Theta}^{(1)},\cdots,\mathbf{\Theta}^{(T-1)}$ can be sampled with evenly spaced time parameters on a Bezier curve in the neural space, thus yielding the following expression
\begin{equation}\label{eq:bezier_geom}
\mathbf{\Theta}^{(t)}=\sum_{i=0}^n\binom{n}{i}\left(\frac{t}{T-1}\right)^i\left(1-\frac{t}{T-1}\right)^{n-i}\mathbf{\Theta}_i,
\end{equation}

\noindent where $\mathbf{\Theta}_0,\mathbf{\Theta}_1,\cdots,\mathbf{\Theta}_n\in\mathbb{R}^{|\mathbf{\Theta}|}$ represent the control points, and $n$ denotes the degree of Bezier curve.

We train the Bezier curve by optimizing all control points simultaneously. The overall loss function is
\begin{equation}\begin{aligned}
\mathcal{L}_F(\mathbf{\Theta}_0,\cdots\mathbf{\Theta}_n)=&\ \mathbb{E}_{t\sim\mathcal{U}(\mathcal{T})}[\mathbb{E}_{\mathbf{x}\sim \mathcal{P}_F^{(t)}}[D_F^{(t)}(\mathbf{x})]]\\
&+\frac{\lambda_F}{|\mathcal{X}^{(0)}|+\cdots+|\mathcal{X}^{(T-1)}|}\sum_{i=0}^n\|\mathbf{\Theta}_i\|_1,
\end{aligned}\end{equation}

\noindent where $\mathcal{T}=\{0,1,2,\cdots,T-1\}$ is all frame indices.

For practical implementation, we rewrite the above loss function as $\mathcal{L}_F(\mathbf{\Theta}_0,\cdots\mathbf{\Theta}_n)=\mathbb{E}_{t\sim\mathcal{U}(\mathcal{T})}[\mathcal{L}_F^{(t)}(\mathbf{\Theta}_0,\cdots\mathbf{\Theta}_n)]$, where
\begin{equation}\label{eq:cnerc3_loss_geom}
\begin{aligned}
\mathcal{L}_F^{(t)}(\mathbf{\Theta}_0,\cdots\mathbf{\Theta}_n)=&\ \mathbb{E}_{\mathbf{x}\sim \mathcal{P}_F^{(t)}}[D_F^{(t)}(\mathbf{x})]\\
&+\frac{\lambda_F}{|\mathcal{X}^{(0)}|+\cdots+|\mathcal{X}^{(T-1)}|}\sum_{i=0}^n\|\mathbf{\Theta}_i\|_1.
\end{aligned}\end{equation}

To average \eqref{eq:cnerc3_loss_geom} over $\mathcal{T}$ with respect to $t$, at each training step, we randomly sample $t\sim\mathcal{U}(\mathcal{T})$. Then we obtain the sample point $\mathbf{\Theta}^{(t)}$ from the trainable control points according to \eqref{eq:bezier_geom}. Finally, we calculate the loss function in \eqref{eq:cnerc3_loss_geom}, and update the control points with a gradient step.

The encoder quantizes and transmits the control points $\mathbf{\Theta}_0,\mathbf{\Theta}_1,\cdots,\mathbf{\Theta}_n$. The decoder receives the quantized control points $\widehat{\mathbf{\Theta}}_0,\widehat{\mathbf{\Theta}}_1,\cdots,\widehat{\mathbf{\Theta}}_n$, and then obtain the lossy sample points $\widehat{\mathbf{\Theta}}^{(0)},\widehat{\mathbf{\Theta}}^{(1)},\cdots,\widehat{\mathbf{\Theta}}^{(T-1)}$ as in \eqref{eq:bezier_geom}. They serve as the parameter sets of $T$ networks that are utilized to reconstruct $T$ frames $\widehat{\mathcal{X}}^{(0)},\widehat{\mathcal{X}}^{(1)},\cdots,\widehat{\mathcal{X}}^{(T-1)}$, separately. The reconstruction process of each frame is the same as in i-NeRC$^3$.

\subsubsection{Attribute Compression}

Let $\bm{\Phi}^{(t)}$ be the parameter set of network $G$ that represents the attributes of the $t$-th frame $C^{(t)}$. The corresponding loss function is $\mathbb{E}_{\widehat{\mathbf{x}}\sim \mathcal{P}_G^{(t)}}[D_G^{(t)}(\widehat{\mathbf{x}})]$.

Similarly to geometry compression, we assume that $\mathbf{\Phi}^{(0)},\mathbf{\Phi}^{(1)},\cdots,\mathbf{\Phi}^{(T-1)}$ can be evenly sampled on a Bezier curve in the neural space $\mathbb{R}^{|\bm{\Phi}|}$, each sample point expressed as
\begin{equation}\label{eq:bezier_attr}
\mathbf{\Phi}^{(t)}=\sum_{i=0}^n\binom{n}{i}\left(\frac{t}{T-1}\right)^i\left(1-\frac{t}{T-1}\right)^{n-i}\mathbf{\Phi}_i,
\end{equation}

\noindent where $\mathbf{\Phi}_0,\mathbf{\Phi}_1,\cdots,\mathbf{\Phi}_n\in\mathbb{R}^{|\mathbf{\Phi}|}$ represent the control points.

The overall loss function for training the control points is
\begin{equation}\begin{aligned}
\mathcal{L}_G(\mathbf{\Phi}_0,\cdots\mathbf{\Phi}_n)=&\ \mathbb{E}_{t\sim\mathcal{U}(\mathcal{T})}[\mathbb{E}_{\widehat{\mathbf{x}}\sim \mathcal{P}_G^{(t)}}[D_G^{(t)}(\widehat{\mathbf{x}})]]\\
&+\frac{\lambda_G}{|\mathcal{X}^{(0)}|+\cdots+|\mathcal{X}^{(T-1)}|}\sum_{i=0}^n\|\mathbf{\Phi}_i\|_1.
\end{aligned}\end{equation}

The training process is similar to that of geometry compression. At each training step, we sample $t\sim\mathcal{U}(\mathcal{T})$, and compute the following loss function for gradient descent.
\begin{equation}\label{eq:cnerc3_loss_attr}
\begin{aligned}
\mathcal{L}_G^{(t)}(\mathbf{\Phi}_0,\cdots\mathbf{\Phi}_n)=&\ \mathbb{E}_{\widehat{\mathbf{x}}\sim \mathcal{P}_G^{(t)}}[D_G^{(t)}(\widehat{\mathbf{x}})]\\
&+\frac{\lambda_G}{|\mathcal{X}^{(0)}|+\cdots+|\mathcal{X}^{(T-1)}|}\sum_{i=0}^n\|\mathbf{\Phi}_i\|_1.
\end{aligned}\end{equation}

After that, the encoder quantizes and transmits the control points. The decoder receives the quantized control points and obtain the sample points as in \eqref{eq:bezier_attr}, which reconstruct the attributes $\widehat{C}^{(0)},\widehat{C}^{(1)},\cdots,\widehat{C}^{(T-1)}$.

\section{Proofs of Propositions}

This section provides the proofs of the two propositions in the main paper. We repeat some definitions here for convenience. The distance between a point $\mathbf{b}$ to a reference point cloud $\mathcal{A}$ is defined as $d(\mathbf{b},\mathcal{A})=\min_{\mathbf{a}\in\mathcal{A}}\|\mathbf{b}-\mathbf{a}\|_2^2$, and the point-to-point (D1) error of a point cloud $\mathcal{B}$ relative to $\mathcal{A}$ is defined as
\begin{equation}
e(\mathcal{B},\mathcal{A})=\frac{1}{|\mathcal{B}|}\sum_{\mathbf{b}\in\mathcal{B}}\min_{\mathbf{a}\in\mathcal{A}}\|\mathbf{b}-\mathbf{a}\|_2^2.
\end{equation}

The D1 PSNR between the original point cloud $\mathcal{X}$ and the reconstructed point cloud $\widehat{\mathcal{X}}$ is
\begin{equation}\label{eq:d1_psnr}
\text{D1 PSNR}=10\log_{10}\frac{3\times(2^N-1)^2}{\max\{e(\widehat{\mathcal{X}},\mathcal{X}),e(\mathcal{X},\widehat{\mathcal{X}})\}}(\text{dB}).
\end{equation}

\subsection{Proof of Proposition 1}

When reconstructing the geometry of a point cloud, we first obtain the OPs $\widehat{p}(\mathbf{x})$ of all voxels in $\mathcal{V}$ by feeding each $\mathbf{x}$ into $F$. It's worth noting that none of these OPs are strictly equal to $0$ or $1$, since the output of $F$ is the sigmoid function. We then sort these OPs in ascending order and remove any duplicate values, yielding a sequence $0<p_1<p_2<\cdots<p_K<1$. These OPs divide the interval $[0,1]$ into $K+1$ subintervals, namely $[0,p_1),[p_1,p_2),\cdots,[p_{K-1},p_K),[p_K,1]$.

Let $\tau_{\max}=p_K$. If $\tau$ lies in the last subinterval, i.e., $\tau\ge \tau_{\max}$, then no voxel has an OP greater than the threshold. In this case, the reconstructed point cloud contains no points, and $D(\tau)$ is undefined. Otherwise, $\tau$ must lie within one of the remaining $K$ subintervals. Consider any such subinterval, which we denote as $[p_{i-1},p_i)$, where $i=1,2,\cdots,K$, and we define $p_0=0$. If $\tau$ lies within $[p_{i-1},p_i)$, then only the voxels with OPs $p_i,p_{i+1},\cdots,p_K$ will be reconstructed as occupied. Therefore, $D(\tau)$ remains constant within this subinterval. Furthermore, $D(\tau)$ is right-continuous, because for any sufficiently small $\delta>0$, such as $\delta=(p_i-\tau)/2$, $\tau+\delta$ still lies within the same subinterval. However, $D(\tau)$ may not be left-continuous at the boundary $\tau=p_{i-1}$, $i=2,3,\cdots,K$, since no matter how small $\delta$ is, $p_{i-1}-\delta$ will always fall into the adjacent subinterval $[p_{i-2},p_{i-1})$, where the voxel(s) with OP $p_{i-1}$ will be additionally reconstructed, potentially altering the value of $D(\tau)$.

\subsection{Proof of Proposition 2}

Based on the proof of Proposition 1, the interval $[0,\tau_{\max})$ is divided into $K$ constant subintervals $[p_0,p_1),[p_1,p_2),\cdots,[p_{K-1},p_K)$, where $p_0=0$ and $p_K=\tau_{\max}$. Let $\widehat{\mathcal{X}}_i$ represent the reconstructed point cloud when $\tau$ is in $[p_{i-1},p_i)$, $i=1,2,\cdots,K$. Clearly, $\widehat{\mathcal{X}}_i$ consists of voxels with OPs $p_i,p_{i+1},\cdots,p_K$. In the following, we consider $e(\widehat{\mathcal{X}}_i,\mathcal{X})$ and $e(\mathcal{X},\widehat{\mathcal{X}}_i)$ as two sequences indexed by $i=1,2,\cdots,K$, and complete our proof in five steps.

\textit{Step 1: Prove that $e(\widehat{\mathcal{X}}_i,\mathcal{X})$ is non-increasing.}

For $i=1,2,,\cdots,K-1$, $\widehat{\mathcal{X}}_i$ can be decomposed into $\widehat{\mathcal{X}}_{i+1}$ and $\delta\widehat{\mathcal{X}}_i=\widehat{\mathcal{X}}_i-\widehat{\mathcal{X}}_{i+1}$, where the former consists of voxels with OPs $p_{i+1},\cdots,p_K$ and the latter consists of voxel(s) with OP $p_i$. According to assumption 2), for any $\mathbf{x}_0\in\delta\widehat{\mathcal{X}}_i$, we have
\begin{equation}
d(\mathbf{x}_0,\mathcal{X})\ge\mathbb{E}_{\mathbf{x}\sim\mathcal{U}(\widehat{\mathcal{X}}_{i+1})}[d(\mathbf{x},\mathcal{X})].
\end{equation}

Next, we take the average of $d(\mathbf{x}_0,\mathcal{X})$ over $\delta\widehat{\mathcal{X}}_i$ w.r.t. $\mathbf{x}_0$, which gives
\begin{equation}
\mathbb{E}_{\mathbf{x}\sim\mathcal{U}(\delta\widehat{\mathcal{X}}_i)}[d(\mathbf{x},\mathcal{X})]\ge\mathbb{E}_{\mathbf{x}\sim\mathcal{U}(\widehat{\mathcal{X}}_{i+1})}[d(\mathbf{x},\mathcal{X})].
\end{equation}

Thus, we can conclude that
\begin{align}
&\mathbb{E}_{\mathbf{x}\sim\mathcal{U}(\widehat{\mathcal{X}}_i)}[d(\mathbf{x},\mathcal{X})]\notag\\
=\ &\frac{|\delta\widehat{\mathcal{X}}_i|}{|\widehat{\mathcal{X}}_i|}\mathbb{E}_{\mathbf{x}\sim\mathcal{U}(\delta\widehat{\mathcal{X}}_i)}[d(\mathbf{x},\mathcal{X})]+\frac{|\widehat{\mathcal{X}}_{i+1}|}{|\widehat{\mathcal{X}}_i|}\mathbb{E}_{\mathbf{x}\sim\mathcal{U}(\widehat{\mathcal{X}}_{i+1})}[d(\mathbf{x},\mathcal{X})]\notag\\
\ge\ &\mathbb{E}_{\mathbf{x}\sim\mathcal{U}(\widehat{\mathcal{X}}_{i+1})}[d(\mathbf{x},\mathcal{X})],
\end{align}

\noindent or equivalently, $e(\widehat{\mathcal{X}}_i,\mathcal{X})\ge e(\widehat{\mathcal{X}}_{i+1},\mathcal{X})$.

\textit{Step 2: Prove that $e(\mathcal{X},\widehat{\mathcal{X}}_i)$ is non-decreasing.}

Similarly to Step 1, we decompose $\widehat{\mathcal{X}}_i$ into $\widehat{\mathcal{X}}_{i+1}$ and $\delta\widehat{\mathcal{X}}_i$, for $i=1,2,\cdots,K-1$. By the definition of $d(\cdot,\cdot)$, for any $\mathbf{x}_0$, we have
\begin{equation}
d(\mathbf{x}_0,\widehat{\mathcal{X}}_i)=\min\{d(\mathbf{x}_0,\delta\widehat{\mathcal{X}}_i),d(\mathbf{x}_0,\widehat{\mathcal{X}}_{i+1})\}\le d(\mathbf{x}_0,\widehat{\mathcal{X}}_{i+1}).
\end{equation}

Averaging over $\mathcal{X}$ for both terms $d(\mathbf{x}_0,\widehat{\mathcal{X}}_i)$ and $d(\mathbf{x}_0,\widehat{\mathcal{X}}_{i+1})$ w.r.t. $\mathbf{x}_0$, we obtain $e(\mathcal{X},\widehat{\mathcal{X}}_i)\le e(\mathcal{X},\widehat{\mathcal{X}}_{i+1})$.

\textit{Step 3: Prove that the minimum value of $e(\widehat{\mathcal{X}}_i,\mathcal{X})$ is $0$.}

Since $e(\widehat{\mathcal{X}}_i,\mathcal{X})$ is non-increasing, its minimum value is achieved when $i=K$. Note that $\widehat{\mathcal{X}}_K$ is exactly the set $\widetilde{\mathcal{X}}_{\max}$ defined in assumption 1).

Suppose $\widehat{\mathcal{X}}_K-\mathcal{X}\ne\varnothing$, where ``$-$" denotes set difference. Then, we must have $\mathcal{X}-\widehat{\mathcal{X}}_K\ne\varnothing$, because by assumption 1), the number of voxels in $\mathcal{X}$ is at least as large as in $\widehat{\mathcal{X}}_K$. Let $\mathbf{x}_1\in\widehat{\mathcal{X}}_K-\mathcal{X}$ and $\mathbf{x}_0\in\mathcal{X}-\widehat{\mathcal{X}}_K$. Since $\mathbf{x}_1\in\widehat{\mathcal{X}}_K$, $\mathbf{x}_1$ has the highest OP, while $\mathbf{x}_0\not\in\widehat{\mathcal{X}}_K$ means that the OP of $\mathbf{x}_0$ is less than $\mathbf{x}_1$. Therefore, we have $\mathbf{x}_1\in\widetilde{\mathcal{X}}_{\rm h}(\mathbf{x}_0)$. Moreover, $x_0\in\mathcal{X}$ implies $d(\mathbf{x}_0,\mathcal{X})=0$, and $\mathbf{x}_1\not\in\mathcal{X}$ implies $d(\mathbf{x}_1,\mathcal{X})>0$, which leads to $\mathbb{E}_{\mathbf{x}\sim\mathcal{U}(\widetilde{\mathcal{X}}_{\rm h}(\mathbf{x}_0))}[d(\mathbf{x},\mathcal{X})]>0$. Thus, we have
\begin{equation}
d(\mathbf{x}_0,\mathcal{X})<\mathbb{E}_{\mathbf{x}\sim\mathcal{U}(\widetilde{\mathcal{X}}_{\rm h}(\mathbf{x}_0))}[d(\mathbf{x},\mathcal{X})],
\end{equation}

\noindent which contradicts assumption 2). Therefore, it must be that $\widehat{\mathcal{X}}_K-\mathcal{X}=\varnothing$, meaning that $\widehat{\mathcal{X}}_K$ is a subset of $\mathcal{X}$. Hence, for any $\mathbf{x}_0\in\widehat{\mathcal{X}}_K$, we have $d(\mathbf{x}_0,\mathcal{X})=0$, and thus $e(\widehat{\mathcal{X}}_K,\mathcal{X})=0$.

\textit{Step 4: Prove that the minimum value of $e(\mathcal{X},\widehat{\mathcal{X}}_i)$ is $0$.}

The minimum value of $e(\mathcal{X},\widehat{\mathcal{X}}_i)$ occurs at $i=1$. By definition, $\widehat{\mathcal{X}}_1=\mathcal{V}$. Clearly, $\mathcal{X}$ is a subset of $\widehat{\mathcal{X}}_1$, so we have $e(\mathcal{X},\widehat{\mathcal{X}}_1)=0$.

\textit{Step 5: Prove that $D(\tau)$ is unimodal.}

From the results of the previous steps, we can conclude that
\begin{gather}
e(\widehat{\mathcal{X}}_K,\mathcal{X})=0=e(\mathcal{X},\widehat{\mathcal{X}}_1)\le e(\mathcal{X},\widehat{\mathcal{X}}_K),\\
e(\widehat{\mathcal{X}}_1,\mathcal{X})\ge e(\widehat{\mathcal{X}}_K,\mathcal{X})=0= e(\mathcal{X},\widehat{\mathcal{X}}_1),
\end{gather}

\noindent or shortly, $e(\widehat{\mathcal{X}}_K,\mathcal{X})\le e(\mathcal{X},\widehat{\mathcal{X}}_K)$ and $e(\widehat{\mathcal{X}}_1,\mathcal{X})\ge e(\mathcal{X},\widehat{\mathcal{X}}_1)$. Therefore, there exists some $I\in\{1,2,\cdots,K-1\}$ such that:
\begin{itemize}
\item For $i=1,\cdots,I$, $e(\widehat{\mathcal{X}}_i,\mathcal{X})\ge e(\mathcal{X},\widehat{\mathcal{X}}_i)$. Hence, $\max\{e(\widehat{\mathcal{X}}_i,\mathcal{X}),e(\mathcal{X},\widehat{\mathcal{X}}_i)\}$ equals $e(\widehat{\mathcal{X}}_i,\mathcal{X})$, which is non-increasing.
\item For $i=I+1,\cdots,K$, $e(\widehat{\mathcal{X}}_i,\mathcal{X})\le e(\mathcal{X},\widehat{\mathcal{X}}_i)$. Hence, $\max\{e(\widehat{\mathcal{X}}_i,\mathcal{X}),e(\mathcal{X},\widehat{\mathcal{X}}_i)\}$ equals $e(\mathcal{X},\widehat{\mathcal{X}}_i)$, which is non-decreasing.
\end{itemize}

According to the definition of D1 PSNR \eqref{eq:d1_psnr}, we have
\begin{equation*}
D(\tau)=10\log_{10}\frac{3\times(2^N-1)^2}{\max\{e(\widehat{\mathcal{X}}_i,\mathcal{X}),e(\mathcal{X},\widehat{\mathcal{X}}_i)\}},\quad\tau\in[p_{i-1},p_i).
\end{equation*}

Let $\tau^*=p_I$. Then $D(\tau)$ is non-decreasing on $[0,\tau^*)$ and non-increasing on $(\tau^*,\tau_{\max})$.


\begin{figure*}[t]
\vspace{-0.1cm}
\centering
\includegraphics[width=2\columnwidth]{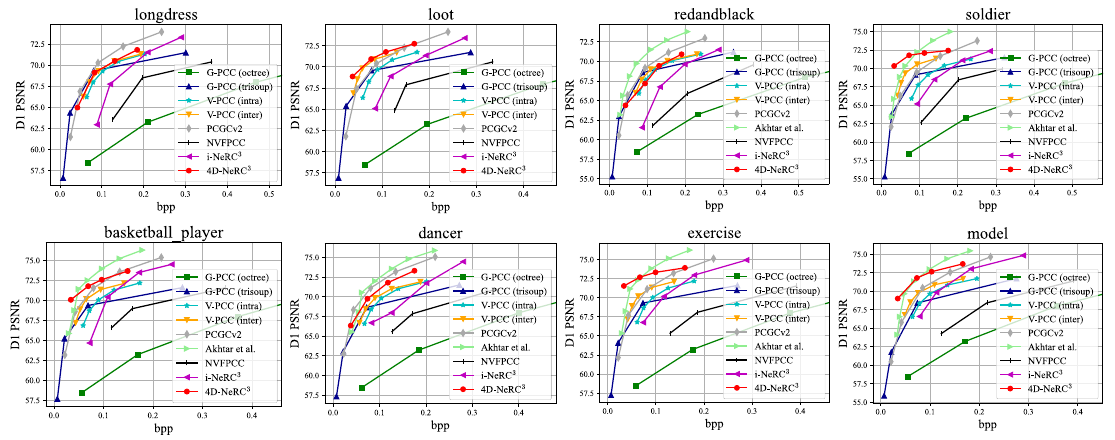}
\vspace{-0.2cm}
\caption{Rate-distortion comparison for geometry compression on the 8iVFB and Owlii datasets. }
\label{fig:_results_geom}
\vspace{-0.1cm}
\end{figure*}

\begin{figure*}[t]
\vspace{-0.1cm}
\centering
\includegraphics[width=2\columnwidth]{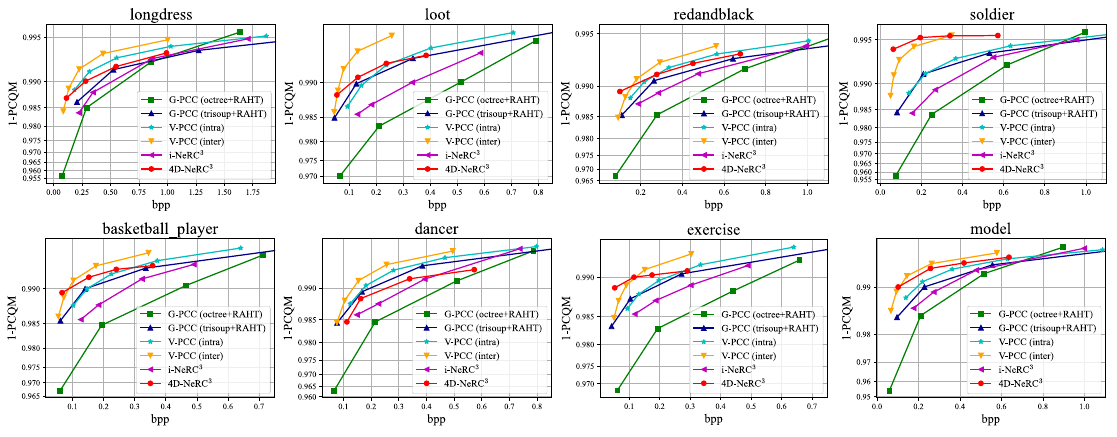}
\vspace{-0.2cm}
\caption{Rate-distortion comparison for joint geometry and attribute compression on the 8iVFB and Owlii datasets. }
\label{fig:_results_joint}
\vspace{-0.1cm}
\end{figure*}

\section{Complexity Analysis}

This section provides a theoretical and experimental analysis on the complexity of the proposed method.

The time complexity of the encoding process primarily stems from network training. We use $T_F$ and $T_G$ to denote the number of training steps for networks $F$ and $G$, and use $|\mathcal{B}|$ to denote the batch size. During geometry compression, we additionally employ golden section search to fine-tune thresholds, which also introduces complexity. We denote the number of fine-tuning steps by $T_{\tau}$. Each fine-tuning step involves forward propagation of network $F$ and D1 PSNR calculation between two point clouds. We assume that nearest neighbor search is implemented by $k$-D trees during D1 PSNR calculation, which yield an average complexity of $O(|\mathcal{V}|\log|\mathcal{V}|)$. Note that in 4D-NeRC$^3$, each group of $T$ frames is represented by a single network, so the network training complexity should be apportioned among $T$ frames. i-NeRC$^3$ can be viewed as a special case of 4D-NeRC$^3$ with $T=1$. Ultimately, the encoding time complexity for geometry compression can be written as
\begin{equation}
C_{\rm geo.enc}=O\left(\frac{1}{T}T_F|\mathcal{B}||\bm{\Theta}|+T_{\tau}(|\mathcal{V}||\bm{\Theta}|+|\mathcal{V}|\log|\mathcal{V}|)\right),
\end{equation}

\noindent while that for attribute compression is
\begin{equation}
C_{\rm att.enc}=O\left(\frac{1}{T}T_G|\mathcal{B}||\bm{\Phi}|\right).
\end{equation}

The decoding time complexity is $C_{\rm geo.dec}=O(|V||\bm{\Theta}|)$ and $C_{\rm att.dec}=O(|\widehat{\mathcal{X}}||\bm{\Phi}|)$ for geometry and attributes, respectively, since the decoding process involves only network inference.

\begin{table}[t]
\setlength\tabcolsep{5pt}
\centering
\caption{Average runtime (seconds per frame) of i-NeRC$^3$ and 4D-NeRC$^3$ on 8iVFB. Different numbers of training steps and values of $T$ are investigated. }
\label{tab:ablation_runtime}
\begin{tabular}{cc|c|cccc}
\toprule
\multicolumn{2}{c|}{\multirow{2}{*}{Scenario}} & \multirow{2}{*}{\# Steps} & i-NeRC$^3$ & \multicolumn{3}{c}{4D-NeRC$^3$} \\
& & & $T=1$ & $T=2$ & $T=4$ & $T=16$ \\
\midrule
\multirow{6}{*}{Geometry} & \multirow{3}{*}{Enc.} & 300K & 2588 & 1572 & 962 & 575 \\
& & 600K & 4951 & 2883 & 1691 & 785 \\
& & 1200K & 9722 & 5365 & 2991 & 1233 \\
& \multirow{3}{*}{Dec.} & 300K & 6.42 & 7.63 & 7.81 & 7.82 \\
& & 600K & 6.43 & 7.81 & 8.03 & 7.96 \\
& & 1200K & 6.47 & 7.49 & 7.64 & 7.76 \\
\midrule
\multirow{6}{*}{Attributes} & \multirow{3}{*}{Enc.} & 200K & 1633 & 873 & 421 & 113 \\
& & 400K & 3199 & 1686 & 868 & 217 \\
& & 800K & 6726 & 3366 & 1757 & 436 \\
& \multirow{3}{*}{Dec.} & 200K & 0.22 & 0.25 & 0.23 & 0.34 \\
& & 400K & 0.21 & 0.24 & 0.24 & 0.36 \\
& & 800K & 0.21 & 0.24 & 0.25 & 0.31 \\
\bottomrule
\end{tabular}
\end{table}

Through the above analysis, the time complexity primarily depends on the number of training steps $T_F,T_G$ and frame group size $T$. Next, we conducted a series of experiments on a platform equipped with AMD EPYC 7402 CPU and NVIDIA GeForce RTX 3090 GPU. Table \ref{tab:ablation_runtime} shows the average runtime of the i-NeRC$^3$ and 4D-NeRC$^3$ on the 8iVFB dataset for different training steps and $T$ values. The experimental results validate our complexity analysis.

\section{Detailed Results}

\begin{figure*}[t]
\vspace{-0.1cm}
\centering
\includegraphics[width=2\columnwidth]{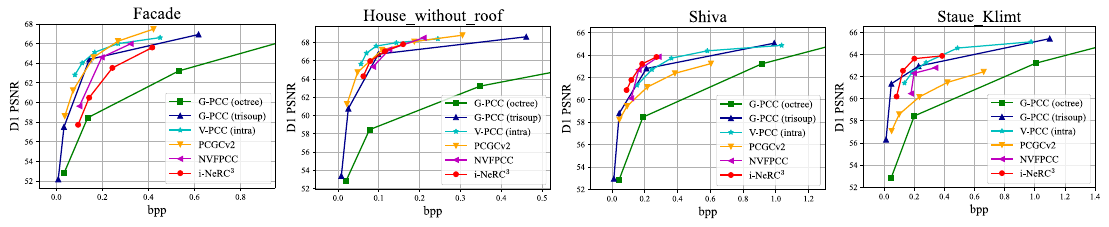}
\vspace{-0.2cm}
\caption{Rate-distortion comparison for geometry compression on the static scenes. }
\label{fig:_results_geom_scenes}
\vspace{-0.1cm}
\end{figure*}

\begin{figure*}[t]
\vspace{-0.1cm}
\centering
\includegraphics[width=2\columnwidth]{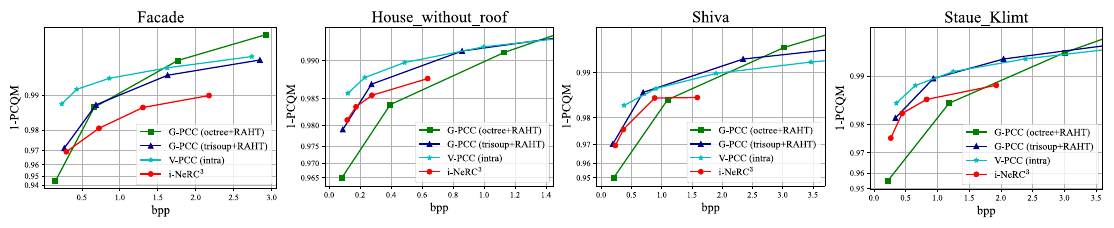}
\vspace{-0.2cm}
\caption{Rate-distortion comparison for joint geometry and attribute compression on the static scenes. }
\label{fig:_results_joint_scenes}
\vspace{-0.1cm}
\end{figure*}

\begin{figure*}[t]
\vspace{-0.1cm}
\centering
\includegraphics[width=2\columnwidth]{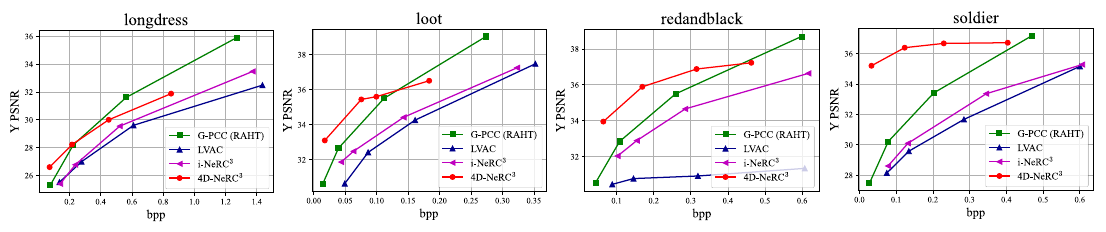}
\vspace{-0.2cm}
\caption{Rate-distortion comparison for attribute compression on the 8iVFB dataset. }
\label{fig:_results_attr}
\vspace{-0.1cm}
\end{figure*}

We present the detailed rate-distortion comparison of each test point cloud sample in this section. The results on 8iVFB and Owlii datasets for geometry compression are shown in Fig. \ref{fig:_results_geom}, while the results for joint compression with attributes are in Fig. \ref{fig:_results_joint}. The results on the four static scenes are in Fig. \ref{fig:_results_geom_scenes} and \ref{fig:_results_joint_scenes}. In Fig. \ref{fig:_results_attr}, we compare our method with LVAC for attribute compression alone.

\begin{figure*}[t]
\vspace{-0.1cm}
\centering
\includegraphics[width=2\columnwidth]{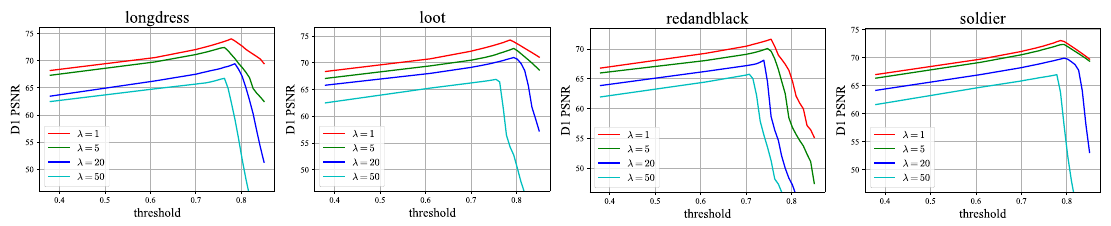}
\vspace{-0.2cm}
\caption{Geometry distortion functions with regard to the threshold. }
\label{fig:_ablation_threshold}
\vspace{-0.1cm}
\end{figure*}

\begin{figure}[t]
\vspace{-0.1cm}
\centering
\includegraphics[width=0.9\columnwidth]{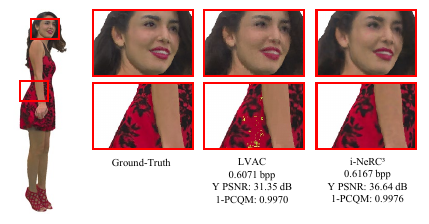}
\vspace{-0.2cm}
\caption{Qualitative visualization of the reconstruction results by LVAC and i-NeRC$^3$ for attribute compression. }
\label{fig:_visual_attr}
\vspace{-0.1cm}
\end{figure}

We observe that LVAC exhibits poor Y PSNR performance on \textit{redandblack}. To investigate the cause, we visualize the reconstruction results of different attribute compression methods in Fig. \ref{fig:_visual_attr}. It is evident that certain points in the reconstruction of LVAC display noticeably erroneous colors, resulting in severe performance degradation.

In Section III-E of the main paper, we demonstrate that the geometry distortion (i.e., D1 PSNR) is a unimodal function of the threshold $\tau$. To assess the impact of this threshold and validate our proposition, we train the network $F$ with each $\lambda$ and then evaluate the reconstruction quality over a range of thresholds. We conduct our experiments on the 8iVFB dataset. The results in Fig. \ref{fig:_ablation_threshold} confirm that the geometry distortion is unimodal.

\section{Open Source}

We will make the source code of NeRC$^3$ publicly available at https://github.com/RhoHenning/NeRC3.


%% file: SecI.tex
Point clouds have become a widely used format for representing 3D objects and scenes across diverse applications such as autonomous driving, augmented reality/virtual reality (AR/VR), digital twins, and robotics \cite{graziosi2020pccstd, wang2021learnedpcgc,shao2025point}. Fundamentally, a point cloud consists of a collection of 3D points distributed throughout volumetric space, each characterized by its spatial coordinates. These coordinates can be quantized into integer values, resulting in the formation of voxels in 3D space, similar to pixels in 2D images -- a process known as voxelization. Beyond geometric data, each point typically carries additional attributes such as color, normal vectors, and reflectance. Recent advances in sensing technologies have enabled the capture of large-scale point clouds with high-resolution spatial and attribute information. Furthermore, real-world objects and dynamic scenes often require capture over time, forming dynamic point cloud sequences. However, the vast data volume, sparse distribution, and unstructured nature of raw point clouds demand substantial memory for storage or high bandwidth for transmission, highlighting the pressing need for efficient point cloud compression (PCC) \cite{bian2024wireless,ruan2024point,shao2024theory}.

The PCC standard was initiated by the Moving Picture Experts Group (MPEG) in 2017 and established two core solutions in 2020 \cite{graziosi2020pccstd}: V-PCC for dynamic point cloud compression and G-PCC for static point clouds and dynamically captured LiDAR sequences. V-PCC operates by transforming 3D point cloud data into 2D projections, particularly suited for dense point clouds with smooth surfaces. These projections can then be efficiently compressed using established image/video codecs like HEVC \cite{sullivan2012hevc}. In contrast, G-PCC directly encodes geometry and attributes within 3D space. For geometry compression, G-PCC employs an octree structure to represent voxelized point cloud geometry, converting it into a binary string for entropy coding. Additionally, G-PCC offers a geometry coding technique called triangle soup (trisoup), which approximates object surfaces using triangular meshes and performs particularly well at low bit rates. For attribute compression, G-PCC employs linear transforms that exploit attribute correlations on the octree structure \cite{liu2021hybrid}, such as the region-adaptive hierarchical transform (RAHT) \cite{de2016raht}.

{
Inspired by the success of learning-based image compression, recent research has explored the potential of deep learning in PCC \cite{liu2019comprehensive, liu2021reduced,liu2021pqanet,liu2022pufagan,liu2024pumask,liu2020model}. These methods typically adopt an autoencoder architecture, where the encoder transforms the input point cloud into latent features for quantization and entropy coding, while the decoder reconstructs the input. Although such learning-based methods outperform traditional approaches, they often require massive point cloud datasets and substantial computational resources for network training, with training processes potentially taking dozens of hours \cite{zhang2024deeppcc}. Furthermore, as highlighted in \cite{alexiou2020unified}, the performance of these networks on unseen test data is significantly influenced by the selection of training datasets.

In recent years, deep neural networks (DNNs) have been employed to implicitly represent 3D objects and scenes by learning continuous functions that take spatial coordinates as inputs and generate corresponding features. This approach, known as implicit neural representations (INRs), has found applications across various research areas, including 3D shape modeling, differentiable rendering, and image/video compression \cite{wu2025lotterycodec}. INRs open up a promising new direction for PCC, as demonstrated by several existing studies \cite{hu2022nvfpcc, isik2022lvac}. However, these studies primarily focus on either geometry or attributes, and the application of INRs to dynamic point cloud compression remains an area with limited exploration.}

In this paper, we present a novel framework for compressing both the geometry and attributes of a single point cloud, named implicit Neural Representations for Colored point Cloud Compression (NeRC$^3$). Drawing inspiration from INRs for 3D shapes, our approach employs two coordinate-based DNNs to implicitly represent { the geometry and attributes of a voxelized point cloud, respectively. The network representing geometry} predicts the occupancy of a given voxel based on its spatial coordinates, outputting an occupancy probability (OP). To address the sparsity of 3D point clouds, we partition the space into smaller cubes, using only voxels within non-empty cubes as network inputs. After calculating the occupancy probability, a threshold is applied to determine binary occupancy, enabling the reconstruction of point cloud geometry by identifying occupied voxels. { The other network, which represents the attributes, generates the attribute values of these occupied voxels based on spatial coordinates as input.
The continuous nature of INRs makes them suited for encoding dense point clouds with smooth features. The encoder trains these two networks specifically for a given point cloud, then quantizes and encodes their parameters, alongside auxiliary information such as non-empty cubes and the occupancy threshold. The decoder retrieves the two networks, which are then used to reconstruct the point cloud's geometry and attributes, respectively.}

For dynamic PCC, our proposed method can be directly applied as an intra-frame compression technique, compressing each frame individually. This approach is termed i-NeRC$^3$. { To further reduce temporal redundancy across frames, we extend our method by exploring additional strategies. Unlike existing approaches that leverage parameter correlations within the neural space to reduce inter-frame redundancy, we introduce 4D spatio-temporal representations (4D-NeRC$^3$) tailored for dynamic point clouds. This method treats point cloud sequences as a single 4D structure, implicitly represented by two DNNs, thereby addressing temporal redundancy directly within the point cloud space.}

Experimental results validate the effectiveness of our proposed methods. { For static PCC, NeRC$^3$ achieves significant performance improvements over octree-based methods employed by the latest G-PCC standard for both geometry compression and joint geometry and attribute compression. It also leads existing INR-based PCC solutions in both geometry compression and attribute compression. For dynamic PCC, 4D-NeRC$^3$ effectively reduces temporal redundancy, outperforming the latest G-PCC and V-PCC standards in standalone geometry compression and demonstrating comparable performance to state-of-the-art learning-based geometry compression schemes. Its joint geometry-attribute compression performance also rivals the latest standards.} Furthermore, our method exhibits superior qualitative evaluation results, generating visually appealing reconstructions rich in detail.

The remainder of this paper is organized as follows. { Section \ref{sec:II} provides a review of existing studies on learning-based and INR-based compression.} Our proposed PCC framework is detailed in Section \ref{sec:III}, with an extension to dynamic PCC described in Section \ref{sec:IV}. Experimental results are presented in Section \ref{sec:V}. Section \ref{sec:VI} concludes this paper.

%% file: SecII.tex
\begin{figure*}[t]
\vspace{-0.1cm}
\centering
\includegraphics[width=2\columnwidth]{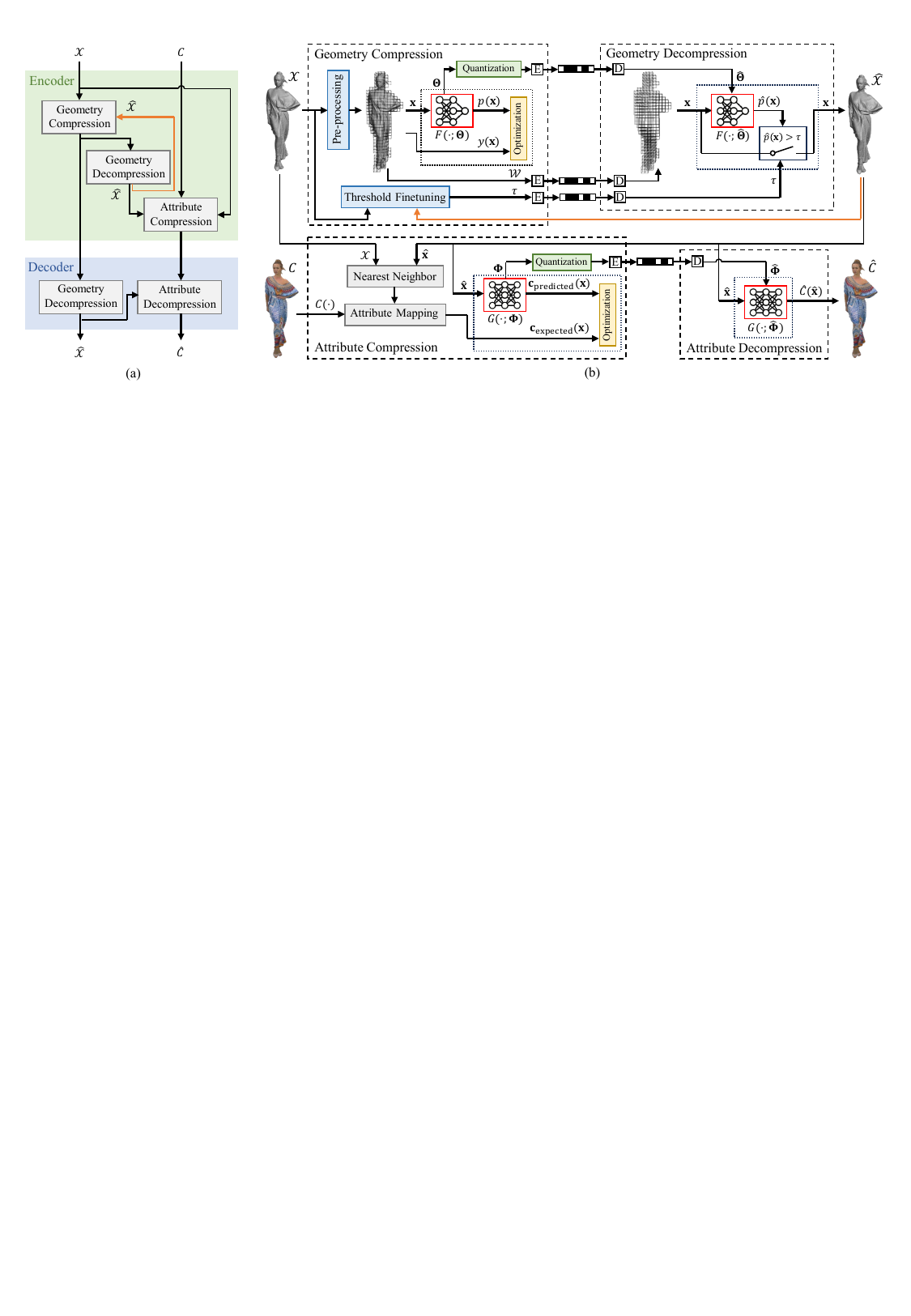}
\vspace{-0.2cm}
\caption{{ (a) A general diagram of point cloud codecs, where attribute compression and decompression is conditioned on the reconstructed geometry $\widehat{\mathcal{X}}$.} (b) Our NeRC$^3$ framework. The encoder optimizes two neural networks $F$ and $G$ to implicitly represent the geometry and attributes, respectively, followed by quantization and encoding of network parameters $\mathbf{\Theta}$ and $\mathbf{\Phi}$. These parameters, along with some auxiliary information (i.e., non-empty cubes $\mathcal{W}$ and the threshold $\tau$), are transmitted to the decoder. The decoder reconstructs a lossy version of the original point cloud using the quantized network parameters and auxiliary information. ``E" and ``D" denote lossless encoding and decoding, respectively. The orange arrow forms a loop for geometry compression, aiming to find the optimal threshold, which will be discussed in Section \ref{sec:threshold}. }
\label{fig:overview}
\vspace{-0.1cm}
\end{figure*}

This section overviews relevant studies on image/video compression and point cloud compression, covering both learning-based and INR-based approaches.

\subsection{Learning-based Compression}

\subsubsection{Image/Video Compression}

Deep learning was initially applied to image compression, demonstrating superior performance compared to traditional methods. Ballé et al. \cite{balle2017learnedic} enhanced rate-distortion performance by jointly optimizing a convolutional autoencoder with an entropy model. This architecture was further extended in \cite{balle2018hyper} by introducing a hyperprior model to achieve more accurate entropy estimation. Such image compression methods can be readily integrated into traditional video compression pipelines by simply replacing key components with DNNs, enabling the construction of learning-based video compression frameworks \cite{lu2019dvc}.

\subsubsection{Point Cloud Compression}

Inspired by these advances, a series of studies have explored the application of deep learning in PCC. Given the lack of structured nature in point clouds compared to images, diverse network architectures have been developed. Point-based methods employ PointNet-style architectures to encode geometry \cite{huang20193dpcgc} or attributes \cite{sheng2021deeppcac} by processing point features, while voxel-based approaches \cite{wang2021learnedpcgc, alexiou2020unified} utilize 3D convolutions to process voxel grids. Given the immense complexity of voxel representations, recent studies \cite{wang2021pcgcv2, liu2022pcgformer,wang2022sparsepcac,zhang2024deeppcc} have adopted sparse tensor representations and employed sparse convolutions to reduce computational costs. Furthermore, multi-scale frameworks in \cite{wang2022sparsepcgc,wang2024unicorn1}, which simultaneously leverage intra-scale and cross-scale correlations, have demonstrated leading performance.

Other studies have focused on utilizing DNNs to refine the coarse reconstruction results of G-PCC. For instance, \cite{zhang2023scalablepcac,zhang2023yoga,fang20223dac} integrated traditional G-PCC into their architectures and enhanced it using sparse convolutions or multilayer perceptrons (MLPs). Additionally, various post-processing models \cite{liu2023grnet,ding2022carnet,xing2023gqenet,xing2024small,wei2025high} have been designed to improve the reconstruction quality of G-PCC.

Recently, deep learning techniques have also been applied to dynamic PCC. Studies such as \cite{fan2022ddpcc,jiang2023ddpcc,akhtar2024ddpcc} employ 3D sparse convolutions to achieve inter-frame prediction within feature space, while Pan et al. \cite{pan2024patchdpcc} introduce point-based compression modules to compress point-wise features. However, these methods are primarily limited to geometry compression, leaving attribute compression insufficiently addressed.

\subsection{INR-based Compression}

Implicit neural representations (a.k.a. neural fields) have emerged as a powerful data representation method capable of handling diverse data types such as audio signals, images, and volumetric content. This approach utilizes neural networks to map spatial or temporal coordinates to corresponding features, with each network typically overfitted to a specific data instance. For instance, neural networks trained on signed distance functions (SDFs) \cite{park2019deepsdf, jiang2020ligr} or occupancy functions \cite{mescheder2019occnet, chen2019imnet} can implicitly represent the geometry of 3D shapes, while images can be represented by neural networks mapping pixel coordinates to RGB values.

\subsubsection{Image/Video Compression}

Thanks to their efficient representational capabilities, INRs have been applied in image and video compression. Dupont et al. \cite{dupont2021coin} first proposed an INR-based image compression method that trains an MLP to represent an image, followed by quantizing and storing the MLP parameters. To simultaneously enhance encoding speed and compression efficiency, \cite{dupont2022coinpp,wu2025mimo} introduced a meta-learning mechanism, where a meta-learned base network is shared and encoding is required only for applied modulation parameters. Similarly, \cite{strumpler2022inric,schwarz2022mscn} proposed to meta-learn an initialization of network parameters, encoding only the parameter updates.

As for video compression, Zhang et al. \cite{zhang2021ipf} followed the traditional video compression pipeline, implementing key components through INRs. In contrast, Chen et al. \cite{chen2021nerv} proposed a novel neural representation for videos (NeRV) that integrates temporal coordinates into INRs. This approach abandons pixel-wise representations, using only frame indices as input to generate complete images through convolutional networks. NeRV's performance was further enhanced through hierarchical encoding \cite{kwan2023hinerv} or combining implicit and explicit approaches \cite{chen2023hnerv}. Differing from pixel-wise or frame-wise methods, Maiya et al. \cite{maiya2023nirvana} proposed a patch-wise framework, where each frame group is assigned with independent networks to generate patch volumes. The network parameters are encoded as residuals relative to the preceding group.

\subsubsection{Point Cloud Compression}

The applications of INRs in image/video compression and 3D shape modeling have laid the foundation for INR-based PCC. Although this field remains nascent, several notable works have emerged. For instance, Hu et al. \cite{hu2022nvfpcc} trained a convolutional neural network alongside input latent codes, where each latent code is utilized by the network to reconstruct occupancies of points within a local region. This approach jointly encodes the network and all latent codes into a representation of the point cloud geometry. In contrast, Pistilli et al. \cite{pistilli2022nic} employ a single coordinate-based neural network to generate point cloud attributes, while LVAC \cite{isik2022lvac} predicts attributes via a coordinate-based network and utilizes input latent vectors as local parameters for enhancement. NeRI \cite{xue2024neri} encodes LiDAR point cloud sequences via INRs by first projecting LiDAR data into a sequence of 2D range images and then applying INR-based image/video compression techniques. However, this approach relies on inherent parameters of the LiDAR sensor and cannot be applied to other point cloud types.


%% file: SecIII.tex
This section present an overview of our proposed framework, as shown in Fig. \ref{fig:overview}, followed by an in-depth discussion of its key components. For now, we concentrate on the compression of static point clouds, with the exploration of dynamic PCC deferred to Section \ref{sec:IV}.

\subsection{Pre-Processing}

We denote the original point cloud at the encoder as $\{\mathcal{X},C\}$, where $\mathcal{X}$ contains the coordinates of all points, representing the geometry, and $C(\cdot)$ maps each point $\mathbf{x}\in\mathcal{X}$ to its corresponding attributes $C(\mathbf{x})$, e.g., RGB colors in this work. The lossy version of the point cloud reconstructed by the decoder is correspondingly denoted as $\{\widehat{\mathcal{X}},\widehat{C}\}$.

\begin{figure}
\vspace{-0.1cm}
\centering
\includegraphics[width=0.9\columnwidth]{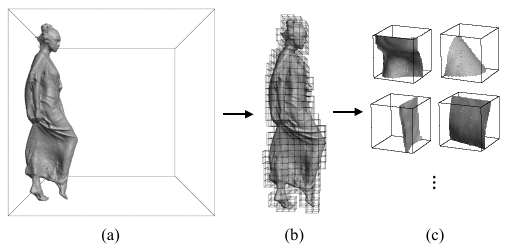}
\vspace{-0.2cm}
\caption{ Pre-processing of point cloud geometry. (a) The entire volumetric space, where most regions are empty. (b) Non-empty cubes containing occupied voxels from the original point cloud. (c) Visualization of each non-empty cube. }
\label{fig:preprocessing}
\vspace{-0.1cm}
\end{figure}

We assume that the point cloud is voxelized with an $N$-bit resolution, meaning all point coordinates are quantized to $N$-bit integers. This voxelization process generates $2^N\times 2^N\times 2^N$ voxels within the volumetric space, each represented by coordinates $\mathbf{x}\in\{0,1,\cdots,2^N-1\}^3$. We denote the entire voxelized space as $\mathcal{S}=\{0,1,\cdots,2^N-1\}^3$. Reconstructing the point cloud requires determining whether each voxel is occupied by a point. However, the vast number of voxels in the space poses a challenge. { As shown in Fig. \ref{fig:preprocessing}(a),} most voxels are empty, making processing all voxels inefficient and extremely time-consuming. To mitigate this issue, we partition the space into $2^M\times 2^M\times 2^M$ cubes, each containing $2^{N-M}\times 2^{N-M}\times 2^{N-M}$ voxels. The position of each cube is represented by a 3D coordinate $\mathbf{w}\in\{0,1,\cdots,2^M-1\}^3$. For any voxel $\mathbf{x}$, the cube containing it can be determined by $\mathbf{w}=\lfloor\mathbf{x}/2^{N-M}\rfloor$, where $\lfloor\cdot\rfloor$ denotes the floor function. { Fig. \ref{fig:preprocessing}(b) illustrates the non-empty cubes, each visualized in Fig. \ref{fig:preprocessing}(c).} The set of coordinates representing all non-empty cubes is denoted by $\mathcal{W}$, expressed as
\begin{equation}
\mathcal{W}=\{\mathbf{w}:\mathbf{w}=\lfloor\mathbf{x}/2^{N-M}\rfloor,\mathbf{x}\in\mathcal{X}\}.
\end{equation}

Since voxels outside these cubes are empty, we only consider voxels within these cubes, the set of which is denoted by
\begin{equation}
\mathcal{V}=\{\mathbf{x}:\lfloor\mathbf{x}/2^{N-M}\rfloor\in\mathcal{W},\mathbf{x}\in\mathcal{S}\}.
\end{equation}

When selecting the value of $M$, a trade-off must be made between the sizes of $\mathcal{W}$ and $\mathcal{V}$. To enable the decoder to recognize non-empty cubes, $\mathcal{W}$ should be encoded as auxiliary information and transmitted to the decoder. Therefore, we choose $M$ to ensure that $\mathcal{W}$ occupies only a small fraction of the total encoded bits. This may result in a large size of $\mathcal{V}$, which consumes significant
memory for storage especially for high-resolution point clouds. However, $\mathcal{V}$ is fundamentally composed of stacked cubes, exhibiting regular hierarchical structures. Consequently, we can access voxels within $\mathcal{V}$ more elegantly without storing all voxels in advance. We denote the set of all local coordinates within a cube as $\mathcal{S}_{\rm local}=\{0,1,\cdots,2^{N-M}-1\}^3$. When sampling a voxel from $\mathcal{V}$, we first sample a cube $\mathbf{w}$ from $\mathcal{W}$, and then sample a local coordinate $\mathbf{x}_{\rm local}$ from $\mathcal{S}_{\rm local}$, where the latter can be generated using random numbers. The global coordinates can then be expressed as $\mathbf{x} = \mathbf{x}_{\rm local}+2^{N-M}\mathbf{w}$. Similarly, traversing voxels in $\mathcal{V}$ requires first traversing all cubes in $\mathcal{W}$ and then enumerating the local coordinates within each cube individually.

\subsection{Compression Pipeline}

\subsubsection{Geometry Compression}

We employ a neural network $F$ parameterized by $\mathbf{\Theta}$ to represent the geometry of the original point cloud by classifying each voxel as occupied or not. The network takes coordinates $\mathbf{x}$ as input and outputs a single real number between $0$ and $1$. As previously stated, we only feed voxels from $\mathcal{V}$ into the network. The output value can be interpreted as the occupancy probability (OP) $p(\mathbf{x})=F(\mathbf{x};\mathbf{\Theta})$.

When training $F$, we optimize the network parameters to ensure that the output probability $p(\mathbf{x})$ approximates the ground-truth occupancy $y(\mathbf{x})$. We label occupied voxels as $1$ and empty voxels as $0$. Thus, the occupancy can be represented by the indicator function $\mathbb{I}(\cdot)$ as
\begin{equation}
y(\mathbf{x})=\mathbb{I}(\mathbf{x}\in\mathcal{X})=\begin{cases}
1, & \mathbf{x}\in\mathcal{X}, \\
0, & \mathbf{x}\not\in\mathcal{X}.
\end{cases}
\end{equation}

The network parameters, serving as an implicit representation of point cloud geometry, undergoes quantization first, followed by lossless entropy coding, and are ultimately transmitted to the decoder. Consequently, the decoder can only access the quantized version of the parameters. For a given step size $\Delta_F$, the quantized parameters can be expressed as $\widehat{\mathbf{\Theta}}=\lfloor\mathbf{\Theta}/\Delta_F\rceil\cdot\Delta_F$, where $\lfloor\cdot\rceil$ denotes the rounding operation. We denote the OP obtained from the quantized parameters as $\widehat{p}(\mathbf{x})=F(\mathbf{x};\widehat{\mathbf{\Theta}})$. After obtaining $\widehat{p}(\mathbf{x})$, a threshold $\tau$ must be selected to determine the binary occupancy. The selection method for $\tau$ will be discussed in Section \ref{sec:threshold}. If the OP $\widehat{p}(\mathbf{x})$ of an input voxel exceeds the threshold $\tau$, the voxel is deemed occupied; otherwise, it is considered empty. All occupied voxels within the non-empty cubes are aggregated to form the reconstructed geometry:
\begin{equation}\label{eq:reconstruct_geometry}
\widehat{\mathcal{X}}=\{\mathbf{x}:F(\mathbf{x};\widehat{\mathbf{\Theta}})>\tau,\mathbf{x}\in\mathcal{V}\}.
\end{equation}

In addition to the parameters $\widehat{\mathbf{\Theta}}$, geometry reconstruction requires the input voxel set $\mathcal{V}$ and the threshold $\tau$, where $\mathcal{V}$ can be determined by $\mathcal{W}$. Therefore, we encode $\mathcal{W}$ and $\tau$ as auxiliary information, transmitting them alongside $\widehat{\mathbf{\Theta}}$ to constitute the complete representation of geometry.

\subsubsection{Attribute Compression}

Before attribute compression, the encoder first performs compression and decompression on the geometry and utilize the reconstructed geometry $\widehat{\mathcal{X}}$ as prior knowledge. We employ another neural network $G$, parameterized by $\mathbf{\Phi}$, to represent the attributes of the point cloud. The network takes occupied voxels $\widehat{\mathbf{x}}$ from the reconstructed geometry as input and generates their RGB colors as output, expressed as $\mathbf{c}_{\rm predicted}(\widehat{\mathbf{x}})=G(\widehat{\mathbf{x}};\mathbf{\Phi})$.

During the training process of $G$, we optimize the network parameters to ensure that the predicted color for each input voxel approximates its expected color. Since the reconstructed geometry $\widehat{\mathcal{X}}$ may differ from the ground-truth geometry $\mathcal{X}$, to minimize the overall distortion, we define the color of each voxel as the ground-truth color of its nearest neighbor in the original point cloud, formulated as
\begin{equation}
\mathbf{c}_{\rm expected}(\widehat{\mathbf{x}})=C\left(\mathop{\arg\min}\limits_{\mathbf{x}\in\mathcal{X}}\|\mathbf{x}-\widehat{\mathbf{x}}\|_2^2\right).
\end{equation}

Similar to geometry compression, the neural field parameters require quantization using a given step size $\Delta_G$. The decoder retrieves only the quantized parameters $\widehat{\mathbf{\Phi}}$, expressed as $\widehat{\mathbf{\Phi}}=\lfloor\mathbf{\Phi}/\Delta_G\rceil\cdot\Delta_G$. Attribute reconstruction is simpler than geometry reconstruction. The coordinates of occupied voxels in the reconstructed geometry are input into $G$ to generate the corresponding attributes, requiring no auxiliary information.
\begin{equation}\label{eq:reconstruct_attributes}
\widehat{C}(\widehat{\mathbf{x}})=G(\widehat{\mathbf{x}};\widehat{\mathbf{\Phi}}).
\end{equation}

\begin{figure}
\vspace{-0.1cm}
\centering
\includegraphics[width=\columnwidth]{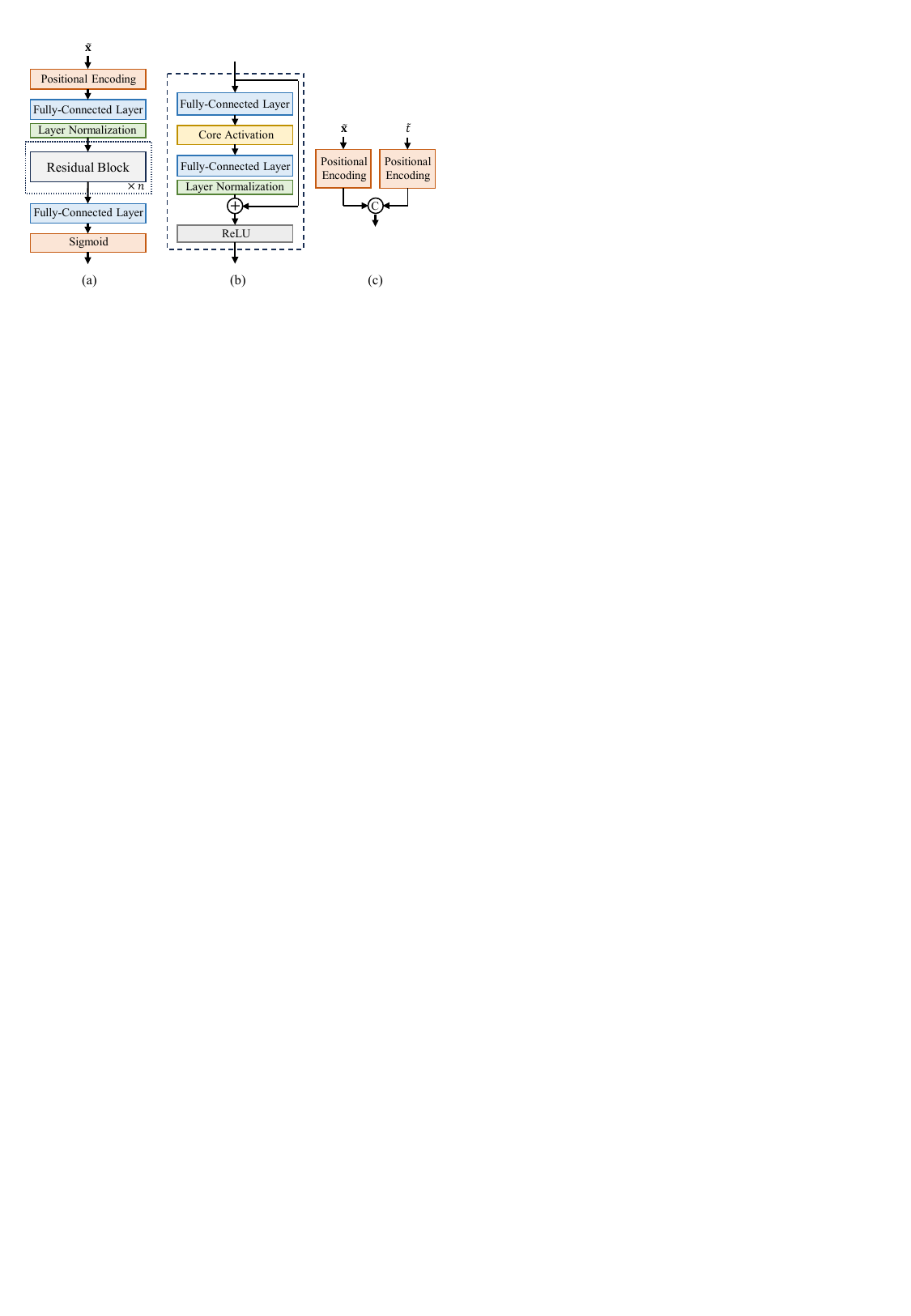}
\vspace{-0.2cm}
\caption{(a) Network structure comprising multiple residual blocks. (b) Detailed structure of each residual block. The plus sign ``$+$" denotes residual connection. (c) Separate positional encoding for the spatial and temporal coordinates. ``C" denotes concatenation. Please refer to Section \ref{sec:4dpc} for details.}
\label{fig:network}
\vspace{-0.1cm}
\end{figure}

\subsection{Network Settings}

\subsubsection{Network Structure}

Both $F$ and $G$ are coordinate-based fully-connected networks that take 3D vectors as inputs. The detailed structure of both networks is presented in Fig. \ref{fig:network}(a)-(b). Each network comprises multiple residual blocks \cite{he2016residual}. To match the dimensions, two additional fully-connected layers are placed at the input and output of the network separately. Each residual block consists of two fully-connected layers. A core activation is placed between these two layers and a ReLU activation is at the output of the block. Network $F$ uses ReLU for its core activation, while $G$ employs the sine activation function \cite{sitzmann2020siren} with a frequency $\omega_0$. Additionally, we perform layer normalization on some network layers. For the output layer, we apply a sigmoid function to obtain output values in the range $[0,1]$. The output dimensions for $F$ and $G$ are 1 and 3, respectively.

We apply positional encoding to map the input coordinate to a higher dimensional space using the encoding function as in \cite{mildenhall2021nerf}. The encoding function is defined as follows:
\begin{equation}\begin{aligned}
\Gamma(\widetilde{\mathbf{x}})=(\widetilde{\mathbf{x}}, \sin(2^0\pi\widetilde{\mathbf{x}}), \cos(2^0\pi\widetilde{\mathbf{x}}),\ldots,\\
\sin(2^{L-1}\pi\widetilde{\mathbf{x}}), \cos(2^{L-1}\pi\widetilde{\mathbf{x}})),
\end{aligned}\end{equation}
\noindent where $L$ determines the number of different frequencies. Before feeding the coordinate $\mathbf{x}$ into the encoding function, it must first be normalized to $[-1,1]$. The normalized coordinate is denoted as $\widetilde{\mathbf{x}}$. 
The encoding function maps each coordinate component from a single scalar to a vector of length $(2L+1)$, which contains more high-frequency variations than the original input.

\subsubsection{Loss Function}

We optimize the two networks $F$ and $G$ by minimizing the following loss functions separately:
\begin{align}
\mathcal{L}_F(\mathbf{\Theta})&=\mathbb{E}_{\mathbf{x}\sim \mathcal{P}_F}[D_F(\mathbf{x})]+\frac{\lambda_F}{|\mathcal{X}|}\|\mathbf{\Theta}\|_1,\\
\mathcal{L}_G(\mathbf{\Phi})&=\mathbb{E}_{\widehat{\mathbf{x}}\sim \mathcal{P}_G}[D_G(\widehat{\mathbf{x}})]+\frac{\lambda_G}{|\mathcal{X}|}\|\mathbf{\Phi}\|_1,
\end{align}
where $D_F(\cdot)$ and $D_G(\cdot)$ denote the distortion of each voxel, while $\mathcal{P}_F$ and $\mathcal{P}_G$ represent the distributions of training samples, which will be discussed in the next subsection. The first term of each loss function represents the expectation of voxel-wise distortion over the sampled voxels, while the second term applies $\ell_1$-regularization to the network parameters.

We adopt $\alpha$-balanced focal loss \cite{lin2017focal} as the geometry distortion loss. The distortion of each voxel $\mathbf{x}$ is defined as
\begin{equation}\label{eq:geometry_distortion}
D_F(\mathbf{x})=-\widetilde{\alpha}(\mathbf{x})\cdot(1-\widetilde{p}(\mathbf{x}))^{\gamma}\log(\widetilde{p}(\mathbf{x})),
\end{equation}
\noindent where
\begin{align}
\widetilde{\alpha}(\mathbf{x})&=y(\mathbf{x})\cdot\alpha+(1-y(\mathbf{x}))\cdot(1-\alpha),\\
\widetilde{p}(\mathbf{x})&=y(\mathbf{x})\cdot p(\mathbf{x})+(1-y(\mathbf{x}))\cdot(1-p(\mathbf{x})),
\end{align}
\noindent $\alpha\in(0,1)$ and $\gamma\ge 0$ are hyperparameters. 

{ Focal loss is based on the traditional binary cross-entropy (BCE) loss. In addition to the original form of BCE, focal loss introduces a modulating factor $(1-\widetilde{p})^{\gamma}$ to reduce the weights assigned to well-classified voxels, enabling the network to focus more on misclassified voxels. A weighting factor $\widetilde{\alpha}$ is simultaneously introduced to balance the numbers of both classes of voxels.} Following \cite{lin2017focal}, we set $\gamma=2$ and define $\alpha$ as the proportion of empty voxels.

We define the attribute distortion loss per voxel as the MSE of the predicted color and expected color:
\begin{equation}
D_G(\widehat{\mathbf{x}})=\|\mathbf{c}_{\rm predicted}(\widehat{\mathbf{x}})-\mathbf{c}_{\rm expected}(\widehat{\mathbf{x}})\|_2^2.
\end{equation}

Some existing works \cite{hu2022nvfpcc, isik2022lvac} employ entropy models to estimate bit rates and minimize rate-distortion loss functions during training. However, we empirically find that applying entropy models slows down and destabilizes the training process. Inspired by $\ell_1$-regularization, we incorporate the $\ell_1$ norm of the original network parameters into our loss functions. Since $\ell_1$-regularization pushes the network parameters to zero, we can readily obtain sparse networks at low bit rates after entropy coding. By adjusting the regularization strengths $\lambda_F$ and $\lambda_G$, encoding performance at different bit rates can be achieved.

\subsection{Sampling Strategy}
The section investigates the sampling strategy of voxels for training, i.e., the distributions $\mathcal{P}_F$ and $\mathcal{P}_G$. For a better explanation, we define a notation to represent the distribution of uniform sampling. Formally, if $\mathbf{a}$ is uniformly sampled from a given set $\mathcal{A}$, it is denoted as $\mathbf{a}\sim\mathcal{U}(\mathcal{A})$, and the probability mass function (PMF) of $\mathbf{a}$ is given by
\begin{equation}
\mathcal{U}(\mathbf{a};\mathcal{A})=\begin{cases}
1/|\mathcal{A}|,&\mathbf{a}\in\mathcal{A},\\
0,&\mathbf{a}\not\in\mathcal{A}.
\end{cases}
\end{equation}

During training, a straightforward strategy is to uniformly sample voxels from all possible inputs of the network, i.e., $\mathcal{P}_F=\mathcal{U}(\mathcal{V})$ and $\mathcal{P}_G=\mathcal{U}(\widehat{\mathcal{X}})$. In our experiment, uniform sampling is applied only for training $G$, while training $F$ employs a more sophisticated strategy, as detailed below.

When training $F$, we manually control the ratio of the occupied voxels in the training samples. This ratio is represented by a hyperparameter $\beta\in(0,1)$. Recall that the focal loss uses parameter $\alpha$ to balance the samples, where $\alpha$ is the proportion of the empty voxels, hence $\alpha=1-\beta$. Following this sampling strategy, we sample voxels from the occupied space $\mathcal{X}$ with probability $\beta$ and from the empty space $\mathcal{V}-\mathcal{X}$ with probability $\alpha$. Thus, the training sample distribution can be expressed as $\mathcal{P}_F=\beta\cdot\mathcal{U}(\mathcal{X})+\alpha\cdot\mathcal{U}(\mathcal{V}-\mathcal{X})$, which implies that the PMF of an arbitrary voxel $\mathbf{x}$ is $\mathcal{P}_F(\mathbf{x})=\beta\cdot\mathcal{U}(\mathbf{x};\mathcal{X})+\alpha\cdot\mathcal{U}(\mathbf{x};\mathcal{V}-\mathcal{X})$.

To uniformly sample voxels from the empty space $\mathcal{V}-\mathcal{X}$, we have to store all these empty voxels in advance. However, for high-resolution point clouds, storing these voxels consumes significant memory. Therefore, we instead uniformly sample voxels from $\mathcal{V}$ rather than directly from $\mathcal{V}-\mathcal{X}$. Recall that voxels in $\mathcal{V}$ can be sampled hierarchically. We first uniformly sample a cube $\mathbf{w}\sim\mathcal{U}(\mathcal{W})$ and then randomly generate local coordinates $\mathbf{x}_{\rm local}\sim\mathcal{U}(\mathcal{S}_{\rm local})$. The resulting global coordinates $\mathbf{x}=\mathbf{x}_{\rm local}+2^{N-M}\mathbf{w}$ can be regarded as uniformly sampled from $\mathcal{V}$, i.e., $\mathbf{x}\sim\mathcal{U}(\mathcal{V})$, since each voxel in $\mathcal{V}$ has an equal probability of being accessed.

However, note that $\mathcal{V}$ contains occupied voxels. If we continue sampling occupied voxels from $\mathcal{X}$ with probability $\beta$, the overall ratio of occupied voxels in the training samples will exceed $\beta$. Therefore, we modify the sampling strategy by rewriting the sample distribution as
\begin{equation}\label{eq:sample_distribution}
\mathcal{P}_F=\beta^*\cdot\mathcal{U}(\mathcal{X})+\alpha^*\cdot\mathcal{U}(\mathcal{V}),
\end{equation}
\noindent where $\beta^*$ and $\alpha^*$ are the adjusted sampling ratios, given by
\begin{equation}
\beta^*=\frac{\beta-\zeta}{1-\zeta},\quad\alpha^*=\frac{\alpha}{1-\zeta},\quad
\zeta=\frac{|\mathcal{X}|}{|\mathcal{V}|}=\frac{|\mathcal{X}|}{(2^{N-M})^3|\mathcal{W}|}.
\end{equation}

As long as the ratio of occupied voxels in the training samples is no less than the real proportion of occupied voxels in $\mathcal{V}$, i.e., $\beta\ge\zeta$, we can replace sampling from $\mathcal{V}-\mathcal{X}$ with sampling from $\mathcal{V}$, with a small adjustment on the sampling ratios. Since the real proportion $\zeta$ is typically very small, $\beta\ge\zeta$ is easily satisfied in practice.

\begin{figure}
\vspace{-0.1cm}
\centering
\includegraphics[width=\columnwidth]{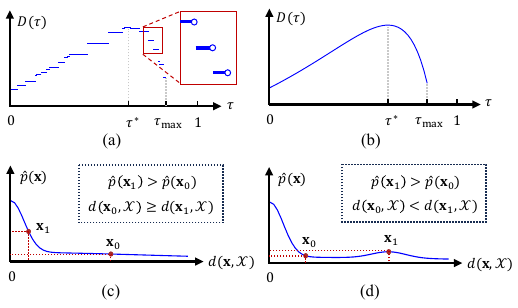}
\vspace{-0.2cm}
\caption{ (a) In theory, $D(\tau)$ is piece-wise constant and right-continuous on $[0,\tau_{\max})$, and is unimodal with its peak at $\tau^*$. (b) In practice, $D(\tau)$ appears like a smooth continuous function. (c) Ideally, voxels with higher OPs should be closer to $\mathcal{X}$. (d) If OPs fluctuate in empty regions, voxels with higher OPs can be farther from $\mathcal{X}$. }
\label{fig:threshold}
\vspace{-0.1cm}
\end{figure}

\subsection{Threshold Fine-tuning}\label{sec:threshold}

As described in \eqref{eq:reconstruct_geometry}, we employ a threshold $\tau$ to predict the occupancy of voxels in $\mathcal{V}$. The value of $\tau$ plays a critical role in the geometry reconstruction process even after the network parameters are optimized and fixed. Consequently, the encoder must fine-tune this threshold to achieve optimal reconstruction quality. It is important to note that the decoder can only access the quantized parameters. To ensure consistency of information between the encoder and decoder, quantization is manually applied to the parameters at the encoder before the threshold fine-tuning step.

We choose point-to-point error (D1) in peak signal-to-noise ratio (PSNR) as the metric to be maximized during threshold fine-tuning process. Formally, let the distance between a point $\mathbf{b}$ to a reference point cloud $\mathcal{A}$ defined as $d(\mathbf{b},\mathcal{A})=\min_{\mathbf{a}\in\mathcal{A}}\|\mathbf{b}-\mathbf{a}\|_2^2$, and the point-to-point error of a point cloud $\mathcal{B}$ relative to $\mathcal{A}$ is defined as
\begin{equation}\label{eq:p2point_error}
e(\mathcal{B},\mathcal{A})=\frac{1}{|\mathcal{B}|}\sum_{\mathbf{b}\in\mathcal{B}}\min_{\mathbf{a}\in\mathcal{A}}\|\mathbf{b}-\mathbf{a}\|_2^2.
\end{equation}

After that, the D1 PSNR between the original point cloud $\mathcal{X}$ and the reconstructed point cloud $\widehat{\mathcal{X}}$ is given by
\begin{equation}\label{eq:d1_psnr}
\text{D1 PSNR}=10\log_{10}\frac{3\times(2^N-1)^2}{\max\{e(\widehat{\mathcal{X}},\mathcal{X}),e(\mathcal{X},\widehat{\mathcal{X}})\}}(\text{dB}).
\end{equation}

We consider the D1 PSNR between $\mathcal{X}$ and $\widehat{\mathcal{X}}$ as a function of the threshold $\tau$, denoted as $D(\tau)$. Consequently, the threshold fine-tuning task can be formulated as finding the maximum point of the function $D(\tau)$. To gain deeper insights into the behavior and characteristics of this function, we propose and analyze several of its properties.

\begin{prop}\label{prop:1}
There exists a threshold boundary $\tau_{\max}\in(0,1)$ such that the following holds:
\begin{itemize}
\item For $\tau\in[0,\tau_{\max})$, $D(\tau)$ is piece-wise constant and right-continuous.
\item For $\tau\in[\tau_{\max},1]$, $D(\tau)$ is undefined.
\end{itemize}
\end{prop}


\begin{prop}\label{prop:2}
 Let the following assumptions hold:
\begin{enumerate}
\item $|\widetilde{\mathcal{X}}_{\max}|\le|\mathcal{X}|$, where $\widetilde{\mathcal{X}}_{\max}=\{\mathbf{x}:\widehat{p}(\mathbf{x})=\tau_{\max},\mathbf{x}\in\mathcal{V}\}$ denotes the set of voxels with the maximum OP.
\item For any $\mathbf{x}_0$, it satisfies
\begin{equation}
d(\mathbf{x}_0,\mathcal{X})\ge\mathbb{E}_{\mathbf{x}\sim\mathcal{U}(\widetilde{\mathcal{X}}_{\rm h}(\mathbf{x}_0))}[d(\mathbf{x},\mathcal{X})],
\end{equation}
\noindent where $\widetilde{\mathcal{X}}_{\rm h}(\mathbf{x}_0)=\{\mathbf{x}:\widehat{p}(\mathbf{x})>\widehat{p}(\mathbf{x}_0),\mathbf{x}\in\mathcal{V}\}$ denotes the set of voxels with OPs greater than that of $\mathbf{x}_0$.
\end{enumerate}
\noindent Then, $D(\tau)$ is unimodal, meaning that there exists a value $\tau^*\in(0,\tau_{\max})$ such that $D(\tau)$ is non-decreasing on $[0,\tau^*)$ and non-increasing on $(\tau^*,\tau_{\max})$.
\end{prop}


{ The proofs of the above propositions are given in the supplementary material.}

Next, we analyze the practical implications of the above propositions within the context of experimental situations. { Fig. \ref{fig:threshold}(a) provides a simplified illustration of $D(\tau)$.} According to the proof of Proposition \ref{prop:1}, the interval $[0,\tau_{\max})$ is segmented into $K$ subintervals such that $D(\tau)$ remains constant within each subinterval, where $K$ is the number of unique values of OPs $\widehat{p}(\mathbf{x})$ for all $\mathbf{x}\in\mathcal{V}$. Given that OPs are predicted by a neural network, it is highly improbable that the network will produce identical output values for different inputs. Therefore, it is reasonable to approximate $K\approx|\mathcal{V}|$. For this reason, it becomes challenging to perceive $D(\tau)$ as a piece-wise constant function in practice. This is because $|\mathcal{V}|$ is typically very large, resulting in each subinterval being exceedingly small and thus indistinguishable. Consequently, $D(\tau)$ appears closer to a smooth continuous function, { as in Fig. \ref{fig:threshold}(b)}.

Regarding Proposition \ref{prop:2}, the low likelihood of duplicate OPs means that assumption 1) generally holds with ease. In theory, assumption 2) can be substituted with a stronger and more intuitive condition: For any $\mathbf{x}_0$ and $\mathbf{x}_1$, where $\mathbf{x}_1$ has a higher OP than $\mathbf{x}_0$, it should hold that $d(\mathbf{x}_0, \mathcal{X}) \ge d(\mathbf{x}_1, \mathcal{X})$. This condition represents an ideal scenario for the distribution of voxel OPs, { as illustrated in Fig. \ref{fig:threshold}(c).} Since the network learns a continuous function fitted to the ground-truth occupancy during training, voxels with higher OPs should theoretically be closer to the occupied voxels in $\mathcal{X}$, resulting in a smaller distance $d(\mathbf{x}, \mathcal{X})$. However, this condition is rarely satisfied in practice. For instance, OPs may exhibit fluctuations in empty regions, { as in Fig. \ref{fig:threshold}(d),} causing voxels with relatively higher OPs to be situated far from $\mathcal{X}$. The assumption 2) in Proposition \ref{prop:2} relaxes this requirement by only necessitating that the average distance of higher-OP voxels remains small, thereby tolerating the presence of outliers with larger distances. { Our experimental results indicate that $D(\tau)$ can be well-approximated as a strictly unimodal function, as shown in the supplementary material, enabling us to employ the golden section search to identify its maximum.}

%% file: SecIV.tex
\begin{figure*}[t]
\vspace{-0.1cm}
\centering
\includegraphics[width=2\columnwidth]{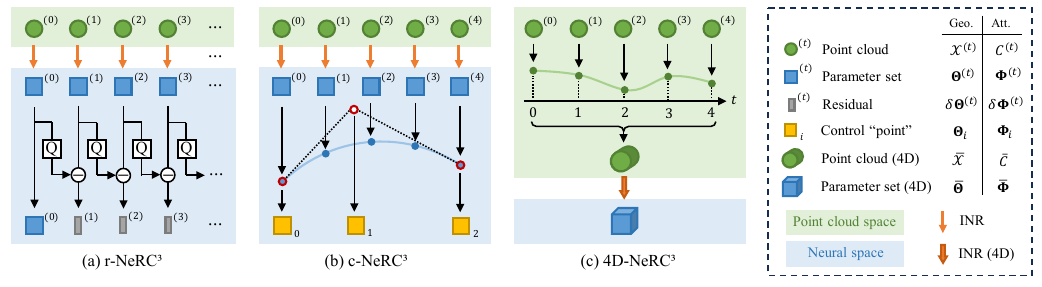}
\vspace{-0.2cm}
\caption{ Diagrams of three extended methods to reduce temporal redundancy. Both (a) r-NeRC$^3$ and (b) c-NeRC$^3$ first transform the geometry/attributes of point clouds into neural space via INR training, with details described in Section \ref{sec:III}. In (a) r-NeRC$^3$, the network parameters in neural space are encoded as residual w.r.t. the quantized parameters of the previous frame. ``Q" denotes quantization. In (b) c-NeRC$^3$, the parameter sets of networks are assumed to lie on a simple curve in neural space. Only the essential control ``points" defining this curve are encoded. (c) 4D-NeRC$^3$ directly eliminates redundancy within the point cloud space by constructing a 4D INR capable of processing multiple frames simultaneously. }
\label{fig:overview_dynamic}
\vspace{-0.1cm}
\end{figure*}

This section explores the problem of dynamic PCC, focusing on the compression of point cloud frame sequences. While NeRC$^3$ can be directly applied as an intra-frame compression method, i.e., compressing each frame independently -- an approach we refer to as i-NeRC$^3$ -- it fails to reduce temporal redundancy between frames. To overcome this limitation, we introduce several methods designed to exploit the temporal correlations, thereby enhancing overall compression efficiency. { Fig. \ref{fig:overview_dynamic} shows the diagrams of these methods.}

\subsection{Residual Compression}

To represent a point cloud sequence, we use a superscript $(\cdot)^{(t)}$ to specify the frame index, where $t=0,1,2,\cdots$. For instance, the $t$-th point cloud frame is denoted as $\{\mathcal{X}^{(t)},C^{(t)}\}$, the corresponding non-empty cubes are denoted as $\mathcal{W}^{(t)}$, and voxels within these cubes are denoted as $\mathcal{V}^{(t)}$.

We first draw our inspiration from a recent study on volumetric video compression \cite{shi2024vvc}. It represented each frame as a 3D scene using NeRF \cite{mildenhall2021nerf} and observed that NeRF parameters of two successive frames share high similarity. Based on this insight, it adopted a strategy of encoding the parameter difference between adjacent frames rather than encoding each frame individually, thereby achieving lower bit rates. We incorporate this strategy into our NeRC$^3$ framework and name it r-NeRC$^3$. { Note that the concept of residual compression originates from traditional image and video coding techniques and has been widely adopted in both video compression and dynamic PCC. Building on this idea of reducing temporal redundancy, we propose r-NeRC$^3$ and use it as a baseline.}

{ Since residual compression is performed in neural space, where both geometry and attributes are represented as network parameters, attributes can be treated on par with geometry. For brevity, here we take geometry compression as an illustrative example. Further details can be found in the supplementary material.} Let $\mathbf{\Theta}^{(t)}$ denote the parameters of the $t$-th network describing the geometry of the $t$-th frame $\mathcal{X}^{(t)}$, where $t=0,1,2,\cdots$. For the first frame, the encoder optimizes parameters $\mathbf{\Theta}^{(0)}$, performs quantization, and then transmits the quantized parameters $\widehat{\mathbf{\Theta}}^{(0)}$ to the decoder, as done in the previous section. For subsequent frames ($t=1,2,\cdots$), the encoder still trains the complete parameters $\mathbf{\Theta}^{(t)}$, but quantizes and transmits only the residual w.r.t. the previous frame, formulated by $\delta\mathbf{\Theta}^{(t)}=\mathbf{\Theta}^{(t)}-\widehat{\mathbf{\Theta}}^{(t-1)}$ and $\delta\widehat{\mathbf{\Theta}}^{(t)}=\lfloor\delta\mathbf{\Theta}^{(t)}/\Delta_F\rceil\cdot\Delta_F$. After receiving $\delta\widehat{\mathbf{\Theta}}^{(t)}$, the decoder retrieves the complete parameters by adding it to the buffered parameters for the previous frame, i.e., $\widehat{\mathbf{\Theta}}^{(t)}=\widehat{\mathbf{\Theta}}^{(t-1)}+\delta\widehat{\mathbf{\Theta}}^{(t)}$. Since the encoder encodes only the residual, we replace the $\ell_1$ norm $\|\mathbf{\Theta}^{(t)}\|_1$ in the loss function with $\|\delta\mathbf{\Theta}^{(t)}\|_1=\|\mathbf{\Theta}^{(t)}-\widehat{\mathbf{\Theta}}^{(t-1)}\|_1$, where $\widehat{\mathbf{\Theta}}^{(t-1)}$ is frozen during the training process.

Residual compression is expected to reduce temporal redundancy, thereby achieving bit rate savings. However, empirical results reveal that its performance actually falls short of compressing each frame individually, particularly for geometry compression. We attribute this to the following reasons. { Unlike \cite{shi2024vvc}, which treats each frame as a solid 3D scene represented by NeRF, our approach focuses on reconstructing the occupancy of points. Since the points primarily concentrate on the scene surfaces rather than the entire solid region, consecutive frames exhibit low similarity in occupancy distribution. Additionally, we employ $\ell_1$-regularization to attain sparse neural networks, a technique not utilized in \cite{shi2024vvc}. Due to the inherent randomness in network optimization, it is challenging to independently train sparse networks with highly similar parameters, resulting in significant differences in network parameters across consecutive frames.}

\begin{figure}
\vspace{-0.1cm}
\centering
\includegraphics[width=\columnwidth]{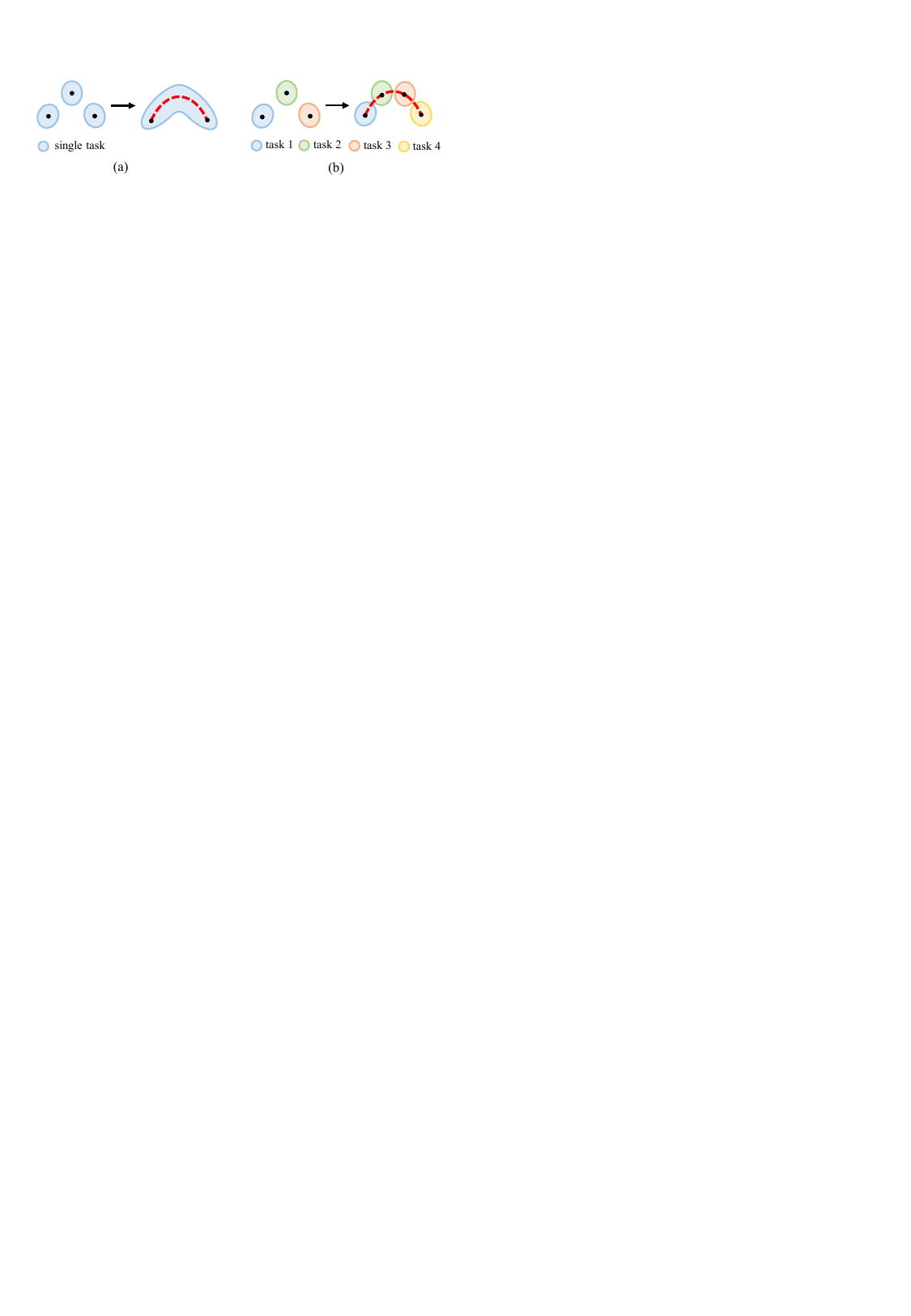}
\vspace{-0.2cm}
\caption{Illustration of the connectivity of optima in neural space, where the colored regions represent low-loss regions for different tasks. (a) A simple curve within one low-loss region connects the optima for a single task. (b) A simple curve traversing the low-loss regions for different tasks connects the optima for these tasks sequentially.}
\label{fig:bezier}
\vspace{-0.1cm}
\end{figure}

\subsection{Connectivity of Optima in Neural Space}

Here, we propose another approach to reduce the temporal redundancy of network parameters. This approach, named as c-NeRC$^3$, draws inspiration from previous studies on mode connectivity \cite{draxler2018dnnmc, garipov2018dnnmc}. These studies state that different optima of neural networks trained on the same task are connected by simple low-loss paths in neural space, such as polygonal chain and Bézier curve. We extend the connectivity of optima to a sequence of tasks by sequentially connecting the optima of different tasks via a single path.

Fig. \ref{fig:bezier} provides a visual illustration. Due to the continuity of loss functions, a specific task corresponds to a connected low-loss region in the neural space. Networks within this region achieve low loss values when evaluated by the corresponding loss function for this task.

\begin{itemize}
\item When only a single task exists, as in Fig. \ref{fig:bezier}(a), the optima of three independently trained networks are isolated on the plane. To find a curve that connects these optima, Garipov et al. \cite{garipov2018dnnmc} first trains two networks separately as the end points, and then trains a simple curve connecting them. The resulting curve lies entirely within a single low-loss region.
\item When four distinct tasks exist, as in Fig. \ref{fig:bezier}(b), the optima of independently trained networks for these tasks are not coplanar in neural space. We directly train a curve traversing the low-loss regions for each task. By appropriately sampling points along this curve, we obtain a point within each low-loss region, representing an optimal neural network for each task.
\end{itemize}

Since our primary focus is on the connectivity of optima in neural space, with slight abuse of notation, this subsection uses the term ``point" to refer to points in neural space, which is essentially the parameter set of a neural network. { Additionally, this subsection uses geometry compression as an illustrative example; further details can be found in the supplementary material.}

We segment a point cloud sequence into groups of size $T$, and encode each group as an independent sequence $\mathcal{X}^{(0)},\mathcal{X}^{(1)},\cdots,\mathcal{X}^{(T-1)}$. For each group, we construct $T$ tasks, where the $t$-th task is to represent $\mathcal{X}^{(t)}$ with a neural network parameterized by $\mathbf{\Theta}^{(t)}$. The corresponding loss function for the $t$-th task is $\mathbb{E}_{\mathbf{x}\sim \mathcal{P}_F^{(t)}}[D_F^{(t)}(\mathbf{x})]$, where $\mathcal{P}_F^{(t)}$ denotes the training sample distribution $\mathcal{P}_F$ in \eqref{eq:sample_distribution} defined on $\mathcal{X}^{(t)}$, and $D_F^{(t)}(\mathbf{x})$ denotes the geometry distortion in \eqref{eq:geometry_distortion} defined on $\mathbf{\Theta}^{(t)}$ and $\mathcal{X}^{(t)}$.

Let $\mathbb{R}^{|\mathbf{\Theta}|}$ denote the neural space, where $|\mathbf{\Theta}|$ is the number of parameters of network $F$. We assume that $\mathbf{\Theta}^{(0)},\mathbf{\Theta}^{(1)},\cdots,\mathbf{\Theta}^{(T-1)}$ can be sampled with evenly spaced time parameters along a Bézier curve in the neural space. Formally, the $t$-th sample point can be expressed as
\begin{equation}\label{eq:bezier}
\mathbf{\Theta}^{(t)}=\sum_{i=0}^n\binom{n}{i}\left(\frac{t}{T-1}\right)^i\left(1-\frac{t}{T-1}\right)^{n-i}\mathbf{\Theta}_i,
\end{equation}
\noindent where $\mathbf{\Theta}_0,\mathbf{\Theta}_1,\cdots,\mathbf{\Theta}_n\in\mathbb{R}^{|\mathbf{\Theta}|}$ represent the control points, and $n$ denotes the degree of Bézier curve. We take the normalized frame index $t/(T-1)\in[0,1]$ as the time parameter of the curve. For brevity, we use $P=n+1$ to denote the number of control points.

We adopt a training process similar to \cite{garipov2018dnnmc}, except that the networks are evaluated by different loss functions and the two end points of the curve are not trained in advance. We train the Bézier curve by optimizing all control points simultaneously. At each training step, we randomly sample a frame index $t$ from all indices in a group, i.e., $t\sim\mathcal{U}(\mathcal{T})$, where $\mathcal{T}=\{0,1,2,\cdots,T-1\}$. Then we obtain the sample point $\mathbf{\Theta}^{(t)}$ from the control points, calculate the loss for the $t$-th task, and update the control points with a gradient step.

The Bézier curve is controlled by only a few control points, yet it can connect more densely distributed points in neural space. In other words, $P$ can be much smaller than $T$. Therefore, the encoder quantizes and transmits the control points $\mathbf{\Theta}_0,\mathbf{\Theta}_1,\cdots,\mathbf{\Theta}_n$ instead of the sample points $\mathbf{\Theta}^{(0)},\mathbf{\Theta}^{(1)},\cdots,\mathbf{\Theta}^{(T-1)}$, thus significantly reducing the bit rates. The overall loss function is formulated as
\begin{equation}\begin{aligned}
\mathcal{L}_F(\mathbf{\Theta}_0,\cdots\mathbf{\Theta}_n)=&\ \mathbb{E}_{t\sim\mathcal{U}(\mathcal{T})}[\mathbb{E}_{\mathbf{x}\sim \mathcal{P}_F^{(t)}}[D_F^{(t)}(\mathbf{x})]]\\
&+\frac{\lambda_F}{|\mathcal{X}^{(0)}|+\cdots+|\mathcal{X}^{(T-1)}|}\sum_{i=0}^n\|\mathbf{\Theta}_i\|_1.
\end{aligned}\end{equation}

The decoder receives the quantized control points $\widehat{\mathbf{\Theta}}_0,\widehat{\mathbf{\Theta}}_1,\cdots,\widehat{\mathbf{\Theta}}_n$, and then obtain the lossy sample points $\widehat{\mathbf{\Theta}}^{(0)},\widehat{\mathbf{\Theta}}^{(1)},\cdots,\widehat{\mathbf{\Theta}}^{(T-1)}$ in the same way as \eqref{eq:bezier}. They serve as the parameters of $T$ networks that are utilized to reconstruct $T$ point cloud frames in a group.

\subsection{4D Spatio-Temporal Representations}\label{sec:4dpc}

The aforementioned methods represent each individual point cloud frame using a neural network, as described in the previous section, and subsequently focus on minimizing temporal redundancy of the network parameters within the neural space. An alternative approach is to directly address redundancy within the point cloud space by constructing a single neural field to simultaneously represent multiple point cloud frames.

Following the last subsection, we take consecutive $T$ frames in the point cloud sequence as a group and encode each group individually. A group of $T$ frames is equivalent to a 4D point cloud, with each point defined by its spatio-temporal coordinate $(t,\mathbf{x})$, where $\mathbf{x}\in\mathcal{X}^{(t)}$ is a point in the $t$-th frame, and $t\in\mathcal{T}$ is the frame index. We define the geometry and attributes of a 4D point cloud as follows:
\begin{gather}
\overline{\mathcal{X}}=\{(t,\mathbf{x}):\mathbf{x}\in\mathcal{X}^{(t)},t\in\mathcal{T}\},\\
\overline{C}(t,\mathbf{x})=C^{(t)}(\mathbf{x}).
\end{gather}

The entire voxelized 4D space contains $T\times 2^N\times 2^N\times 2^N$ 4D voxels. We partition the 4D space into $T\times 2^M\times 2^M\times 2^M$ cubes, each containing $1\times 2^{N-M}\times 2^{N-M}\times 2^{N-M}$ voxels. Note that all voxels in the same cube share one temporal coordinate. The non-empty cubes and the voxels within these cubes in 4D space can be formulated as
\begin{align}
\overline{\mathcal{W}}&=\{(t,\mathbf{w}):\mathbf{w}\in\mathcal{W}^{(t)},t\in\mathcal{T}\},\\
\overline{\mathcal{V}}&=\{(t,\mathbf{x}):\mathbf{x}\in\mathcal{V}^{(t)},t\in\mathcal{T}\}.
\end{align}
Therefore, the INR of a single point cloud can be easily extended to representing several point cloud frames, resulting in much lower bit rates. We name this approach as 4D-NeRC$^3$.

The geometry and attributes of the 4D point cloud are represented by two neural networks $\overline{F}$ and $\overline{G}$ parameterized by $\overline{\mathbf{\Theta}}$ and $\overline{\mathbf{\Phi}}$, respectively. Both networks take spatio-temporal coordinates $(t,\mathbf{x})$ as input. When the temporal coordinate $t$ is fixed, both networks process only the spatial coordinates, thereby representing a single frame and regressing to 3D representations described in the previous section. We follow the network structure depicted in Fig. \ref{fig:network}. Note that positional encoding is applied separately to the spatial coordinate $\mathbf{x}$ and temporal coordinate $t$. We select different values of $L$ for the two encodings, denoted by $L_{\mathbf{x}}$ and $L_t$, and concatenate the resulting vectors before feeding them into the first fully-connected layer, as demonstrated in Fig. \ref{fig:network}(c).

When training $\overline{G}$, we match the expected color of a voxel to its nearest neighbor within the same 3D frame rather than its nearest neighbor in 4D space, since the overall attribute distortion of point clouds is evaluated for each frame individually. Furthermore, instead of utilizing a single threshold for all frames, we assign an independent threshold $\tau^{(t)}$ to each frame for better reconstruction quality. By fixing the temporal coordinate $t$ and following (\ref{eq:reconstruct_geometry}) and (\ref{eq:reconstruct_attributes}), the $t$-th point cloud frame can be reconstructed as
\begin{gather}
\widehat{\mathcal{X}}^{(t)}=\{\mathbf{x}:\overline{F}(t,\mathbf{x};\widehat{\overline{\mathbf{\Theta}}})>\tau^{(t)},\mathbf{x}\in\mathcal{V}^{(t)}\},\\
\widehat{C}^{(t)}(\widehat{\mathbf{x}})=\overline{G}(t,\widehat{\mathbf{x}};\widehat{\overline{\mathbf{\Phi}}}).
\end{gather}

%% file: SecV.tex
\begin{figure}[t]
\vspace{-0.1cm}
\centering
\includegraphics[width=0.9\columnwidth]{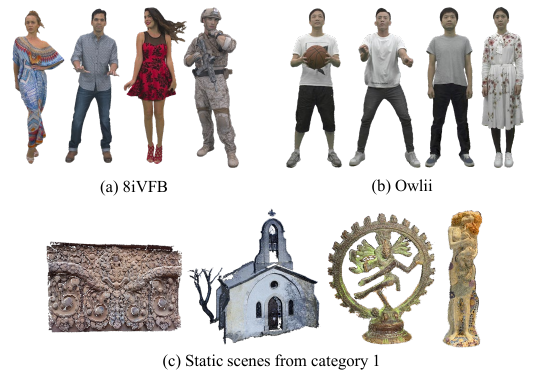}
\vspace{-0.2cm}
\caption{ Visualization of test point clouds.}
\label{fig:dataset}
\vspace{-0.1cm}
\end{figure}

\begin{table*}[t]
\centering
\caption{ BD-PSNR (dB) of different methods against the anchor G-PCC (octree) for geometry compression. }
\label{tab:bdpsnr_geom}
\begin{threeparttable}
\begin{tabular}{c|cccccccccc|cccccc}
\toprule
\multirow{3}{*}{Point Cloud} & \multicolumn{10}{c|}{Static} & \multicolumn{6}{c}{Dynamic} \\
\cmidrule{2-17}
& \multicolumn{2}{c}{G-PCC (trisoup)} & \multicolumn{2}{c}{V-PCC (intra)} & \multicolumn{2}{c}{PCGCv2} & \multicolumn{2}{c}{NVFPCC} & \multicolumn{2}{c|}{i-NeRC$^3$} & \multicolumn{2}{c}{V-PCC (inter)} & \multicolumn{2}{c}{Akhtar et al.} & \multicolumn{2}{c}{4D-NeRC$^3$} \\
& D1 & D2 & D1 & D2 & D1 & D2 & D1 & D2 & D1 & D2 & D1 & D2 & D1 & D2 & D1 & D2 \\
\midrule
\textit{longdress} & 8.98 & 6.77 & 9.02 & 6.83 & 10.56 & 9.28 & 4.93 & 1.92 & 7.43 & 4.83 & 9.43 & 7.35 & - & - & 9.66 & 7.57 \\
\textit{loot} & 9.04 & 6.77 & 9.65 & 7.64 & 10.48 & 9.06 & 6.00 & 3.17 & 7.91 & 5.41 & 11.22 & 9.54 & - & - & 11.30 & 9.62 \\
\textit{red\&black} & 8.54 & 6.15 & 8.35 & 6.12 & 9.80 & 8.67 & 3.11 & -0.41 & 6.15 & 3.30 & 8.75 & 6.60 & 11.55 & 11.00 & 8.43 & 6.06 \\
\textit{soldier} & 8.76 & 6.61 & 8.36 & 6.15 & 10.12 & 8.88 & 4.96 & 1.91 & 7.51 & 4.80 & 10.98 & 9.40 & 12.84 & 12.26 & 11.97 & 10.66 \\
\textbf{Average} & \textbf{8.83} & \textbf{6.58} & \textbf{8.85} & \textbf{6.69} & \textbf{10.24} & \textbf{8.97} & \textbf{4.75} & \textbf{1.65} & \textbf{7.25} & \textbf{4.59} & \textbf{10.10} & \textbf{8.22} & \textbf{-} & \textbf{-} & \textbf{10.34} & \textbf{8.48} \\
\midrule
\textit{basketball} & 8.74 & 7.20 & 9.46 & 7.28 & 11.53 & 9.61 & 5.56 & 2.48 & 9.15 & 6.63 & 10.79 & 8.92 & 13.29 & 11.83 & 12.08 & 10.18 \\
\textit{dancer} & 8.59 & 7.07 & 8.95 & 6.62 & 11.48 & 9.61 & 4.72 & 1.31 & 7.42 & 4.32 & 9.55 & 7.35 & 12.14 & 10.47 & 10.66 & 8.39 \\
\textit{exercise} & 8.79 & 6.91 & 9.47 & 7.21 & 11.24 & 9.25 & 4.21 & 0.87 & 9.01 & 6.42 & 11.11 & 9.23 & 13.62 & 12.15 & 12.84 & 11.16 \\
\textit{model} & 8.19 & 5.89 & 8.69 & 5.67 & 10.27 & 7.84 & 3.46 & -0.22 & 8.80 & 5.95 & 9.83 & 7.09 & 12.35 & 10.15 & 11.94 & 9.52 \\
\textbf{Average} & \textbf{8.58} & \textbf{6.77} & \textbf{9.14} & \textbf{6.70} & \textbf{11.13} & \textbf{9.08} & \textbf{4.49} & \textbf{1.11} & \textbf{8.60} & \textbf{5.83} & \textbf{10.32} & \textbf{8.15} & \textbf{12.85} & \textbf{11.15} & \textbf{11.88} & \textbf{9.81} \\
\midrule
\textit{Facade} & 5.13 & 4.75 & 5.67 & 4.67 & 5.43 & 6.04 & 4.33 & 2.03 & 2.54 & -0.16 & - & - & - & - & - & - \\
\textit{House} & 6.87 & 5.48 & 7.96 & 6.33 & 7.64 & 6.66 & 7.19 & 4.70 & 7.30 & 4.87 & - & - & - & - & - & - \\
\textit{Shiva} & 4.04 & 2.61 & 2.95 & 1.66 & 2.88 & 3.97 & 4.49 & 2.39 & 4.85 & 2.86 & - & - & - & - & - & - \\
\textit{Staue} & 4.64 & 3.11 & 3.63 & 2.29 & 1.89 & 3.09 & 3.91 & 1.62 & 5.06 & 3.15 & - & - & - & - & - & - \\
\textbf{Average} & \textbf{5.17} & \textbf{3.99} & \textbf{5.05} & \textbf{3.74} & \textbf{4.46} & \textbf{4.94} & \textbf{4.98} & \textbf{2.69} & \textbf{4.94} & \textbf{2.68} & \textbf{-} & \textbf{-} & \textbf{-} & \textbf{-} & \textbf{-} & \textbf{-} \\
\bottomrule
\end{tabular}
\begin{tablenotes}
\footnotesize
\item Akhtar et al. trained their model using \textit{longdress} and \textit{loot}. Hence, we evaluate their method on the remaining sequences.
\end{tablenotes}
\end{threeparttable}
\end{table*}

\begin{table}[t]
\centering
\caption{ BD-PSNR (dB) of different methods against the anchor G-PCC (RAHT) for attribute compression.}
\label{tab:bdpsnr_attr}
\begin{tabular}{c|cccc|cc}
\toprule
\multirow{3}{*}{Point Cloud} & \multicolumn{4}{c|}{Static} & \multicolumn{2}{c}{Dynamic} \\
\cmidrule{2-7}
& \multicolumn{2}{c}{LVAC} & \multicolumn{2}{c|}{i-NeRC$^3$} & \multicolumn{2}{c}{4D-NeRC$^3$} \\
& Y & YUV & Y & YUV & Y & YUV \\
\midrule
\textit{longdress} & -2.23 & -2.34 & -1.82 & -1.80 & -0.07 & -0.04 \\
\textit{loot} & -2.47 & -2.44 & -1.74 & -1.72 & 1.21 & 1.23 \\
\textit{red\&black} & -4.35 & -4.40 & -1.22 & -1.28 & 1.51 & 1.51 \\
\textit{soldier} & -2.69 & -2.68 & -1.98 & -1.98 & 4.57 & 4.53 \\
\textbf{Average} & \textbf{-2.94} & \textbf{-2.97} & \textbf{-1.69} & \textbf{-1.70} & \textbf{1.81} & \textbf{1.81} \\
\bottomrule
\end{tabular}
\end{table}

\begin{table}[t]
\centering
\caption{ BD-PCQM ($\times 10^{-3}$) of different methods against the anchor G-PCC (RAHT) for attribute compression.}
\label{tab:bdpcqm_attr}
\begin{tabular}{c|cc|c}
\toprule
\multirow{2}{*}{Point Cloud} & \multicolumn{2}{c|}{Static} & Dynamic \\
\cmidrule{2-4}
& LVAC & i-NeRC$^3$ & 4D-NeRC$^3$ \\
\midrule
\textit{longdress} & -1.15 & -1.39 & 0.53 \\
\textit{loot} & -1.86 & -1.68 & 1.85 \\
\textit{red\&black} & -1.67 & -0.85 & 1.27 \\
\textit{soldier} & -1.56 & -1.62 & 3.18 \\
\textbf{Average} & \textbf{-1.56} & \textbf{-1.39} & \textbf{1.71} \\
\bottomrule
\end{tabular}
\end{table}

\begin{table*}[t]
\setlength\tabcolsep{3.6pt}
\centering
\caption{ BD-PSNR (dB) of different methods against the anchor G-PCC (octree+RAHT) for joint compression. }
\label{tab:bdpsnr_joint}
\begin{tabular}{c|cccccccccccc|cccccccc}
\toprule
\multirow{3}{*}{Point Cloud} & \multicolumn{12}{c|}{Static} & \multicolumn{8}{c}{Dynamic} \\
\cmidrule{2-21}
& \multicolumn{4}{c}{G-PCC (trisoup+RAHT)} & \multicolumn{4}{c}{V-PCC (intra)} & \multicolumn{4}{c|}{i-NeRC$^3$} & \multicolumn{4}{c}{V-PCC (inter)} & \multicolumn{4}{c}{4D-NeRC$^3$} \\
& D1 & D2 & Y & YUV & D1 & D2 & Y & YUV & D1 & D2 & Y & YUV & D1 & D2 & Y & YUV & D1 & D2 & Y & YUV \\
\midrule
\textit{longdress} & -1.65 & -5.60 & 2.33 & 2.34 & 3.16 & 0.35 & 4.33 & 3.97 & 3.42 & 0.68 & 0.51 & 0.49 & 6.04 & 3.52 & 5.52 & 5.26 & 5.12 & 2.32 & 2.48 & 2.47 \\
\textit{loot} & 3.63 & 0.50 & 5.40 & 5.39 & 5.23 & 2.70 & 5.65 & 5.56 & 5.63 & 3.08 & 2.11 & 2.11 & 9.60 & 7.56 & 9.18 & 9.10 & 9.02 & 6.93 & 5.21 & 5.23 \\
\textit{red\&black} & -0.18 & -3.82 & 3.34 & 3.50 & 4.05 & 1.37 & 4.91 & 4.56 & 2.93 & -0.06 & 1.60 & 1.67 & 6.68 & 4.24 & 5.65 & 5.40 & 4.09 & 0.95 & 4.47 & 4.59 \\
\textit{soldier} & 0.85 & -2.56 & 4.64 & 4.62 & 3.67 & 0.94 & 4.79 & 4.71 & 4.10 & 1.19 & 1.65 & 1.64 & 8.92 & 6.87 & 9.90 & 9.80 & 9.08 & 7.26 & 8.79 & 8.74 \\
\textbf{Average} & \textbf{0.66} & \textbf{-2.87} & \textbf{3.93} & \textbf{3.96} & \textbf{4.03} & \textbf{1.34} & \textbf{4.92} & \textbf{4.70} & \textbf{4.02} & \textbf{1.22} & \textbf{1.47} & \textbf{1.48} & \textbf{7.81} & \textbf{5.55} & \textbf{7.56} & \textbf{7.39} & \textbf{6.83} & \textbf{4.37} & \textbf{5.24} & \textbf{5.26} \\
\midrule
\textit{basketball} & 2.59 & -0.18 & 5.14 & 5.09 & 5.35 & 2.82 & 5.77 & 5.58 & 6.95 & 4.39 & 2.91 & 2.85 & 8.65 & 6.40 & 7.83 & 7.67 & 9.44 & 7.04 & 6.62 & 6.53 \\
\textit{dancer} & 1.16 & -1.08 & 4.75 & 4.72 & 4.54 & 1.91 & 5.03 & 4.88 & 4.87 & 1.94 & 2.19 & 2.12 & 7.11 & 4.61 & 6.02 & 5.91 & 6.53 & 3.72 & 3.29 & 3.23 \\
\textit{exercise} & 3.47 & 0.74 & 4.63 & 4.58 & 5.57 & 3.05 & 4.80 & 4.65 & 6.87 & 4.20 & 2.68 & 2.64 & 9.17 & 6.96 & 7.00 & 6.87 & 11.25 & 9.31 & 6.33 & 6.26 \\
\textit{model} & -1.38 & -3.82 & 3.36 & 3.34 & 3.55 & 0.35 & 3.82 & 3.67 & 5.18 & 2.44 & 1.19 & 1.16 & 6.84 & 3.89 & 6.22 & 6.10 & 7.19 & 4.28 & 4.85 & 4.80 \\
\textbf{Average} & \textbf{1.46} & \textbf{-1.09} & \textbf{4.47} & \textbf{4.43} & \textbf{4.75} & \textbf{2.03} & \textbf{4.86} & \textbf{4.70} & \textbf{5.97} & \textbf{3.24} & \textbf{2.24} & \textbf{2.19} & \textbf{7.94} & \textbf{5.47} & \textbf{6.77} & \textbf{6.64} & \textbf{8.60} & \textbf{6.09} & \textbf{5.27} & \textbf{5.21} \\
\midrule
\textit{Facade} & -5.09 & -5.72 & 1.63 & 1.63 & 0.56 & -1.02 & 3.63 & 3.56 & -3.03 & -6.35 & -0.77 & -0.78 & - & - & - & - & - & - & - & - \\
\textit{House} & -1.32 & -2.18 & 1.79 & 1.79 & 3.14 & 1.15 & 1.99 & 1.97 & 4.70 & 1.90 & 1.42 & 1.42 & - & - & - & - & - & - & - & - \\
\textit{Shiva} & -3.59 & -5.41 & 1.16 & 1.17 & -1.30 & -2.72 & 0.69 & 0.68 & 1.21 & -1.14 & 0.88 & 0.86 & - & - & - & - & - & - & - & - \\
\textit{Staue} & -2.40 & -4.74 & 1.69 & 1.74 & -0.49 & -2.05 & 2.15 & 2.11 & 1.42 & -1.06 & 1.23 & 1.26 & - & - & - & - & - & - & - & - \\
\textbf{Average} & \textbf{-3.10} & \textbf{-4.51} & \textbf{1.57} & \textbf{1.58} & \textbf{0.48} & \textbf{-1.16} & \textbf{2.12} & \textbf{2.08} & \textbf{1.08} & \textbf{-1.66} & \textbf{0.69} & \textbf{0.69} & \textbf{-} & \textbf{-} & \textbf{-} & \textbf{-} & \textbf{-} & \textbf{-} & \textbf{-} & \textbf{-} \\
\bottomrule
\end{tabular}
\end{table*}

\begin{table}[t]
\setlength\tabcolsep{5pt}
\centering
\caption{ BD-PCQM ($\times 10^{-3}$) of different methods against the anchor G-PCC (octree+RAHT) for joint compression.}
\label{tab:bdpcqm_joint}
\begin{tabular}{c|ccc|cc}
\toprule
\multirow{2}{*}{Point Cloud} & \multicolumn{3}{c|}{Static} & \multicolumn{2}{c}{Dynamic} \\
\cmidrule{2-6}
& \makecell[c]{G-PCC\\(trisoup\\+RAHT)} & \makecell[c]{V-PCC\\(intra)} & i-NeRC$^3$ & \makecell[c]{V-PCC\\(inter)} & 4D-NeRC$^3$ \\
\midrule
\textit{longdress} & 1.69 & 3.48 & 0.76 & 9.58 & 6.09 \\
\textit{loot} & 8.17 & 7.19 & 3.89 & 14.69 & 10.99 \\
\textit{red\&black} & 4.77 & 4.79 & 2.45 & 8.96 & 7.73 \\
\textit{soldier} & 8.74 & 5.79 & 3.21 & 19.22 & 20.20 \\
\textbf{Average} & \textbf{5.84} & \textbf{5.31} & \textbf{2.58} & \textbf{13.11} & \textbf{11.25} \\
\midrule
\textit{basketball} & 7.19 & 5.56 & 3.39 & 11.24 & 10.07 \\
\textit{dancer} & 6.63 & 4.95 & 2.36 & 9.32 & 4.70 \\
\textit{exercise} & 7.42 & 5.97 & 4.04 & 11.56 & 11.59 \\
\textit{model} & 4.32 & 3.41 & 1.30 & 10.96 & 7.02 \\
\textbf{Average} & \textbf{6.39} & \textbf{4.97} & \textbf{2.77} & \textbf{10.77} & \textbf{8.35} \\
\midrule
\textit{Facade} & 0.27 & 6.43 & -5.94 & - & - \\
\textit{House} & 5.08 & 5.38 & 5.16 & - & - \\
\textit{Shiva} & 7.15 & 3.40 & 6.40 & - & - \\
\textit{Staue} & 5.99 & 6.39 & 7.28 & - & - \\
\textbf{Average} & \textbf{4.62} & \textbf{5.40} & \textbf{3.23} & \textbf{-} & \textbf{-} \\
\bottomrule
\end{tabular}
\end{table}

This section performs numerical experiments to verify the effectiveness of the proposed schemes.

\subsection{Experimental Settings}

\subsubsection{Test Datasets}

We validate our methods using four point cloud sequences from 8i Voxelized Full Bodies \cite{deon20178ivfb} (8iVFB), namely \textit{longdress}, \textit{loot}, \textit{redandblack}, and \textit{soldier}, and four sequences from Owlii Dynamic Human Textured Mesh Sequence Dataset \cite{xu2017owlii} (Owlii), namely \textit{basketball\_player}, \textit{dancer}, \textit{exercise} and \textit{model}. We test the first frame of each sequence to validate static PCC performance and test the first 32 frames of each sequence to evaluate dynamic PCC performance. { Additionally, we employ four static point clouds, namely \textit{Facade}, \textit{House\_without\_roof}, \textit{Shiva}, and \textit{Staue\_Klimt}, to enrich test data for static PCC. These four static scenes are selected from category 1 of test samples from the common test conditions \cite{schwarz2018ctc}. All test point clouds are visualized in Fig. \ref{fig:dataset}. To reduce computational complexity and mitigate sparsity in low-density point clouds, we voxelize all point clouds with a 10-bit resolution.}

\subsubsection{Parameter Settings}

The 3D space of each frame is partitioned into $32\times 32\times 32$ cubes, i.e., $M=5$. { Network $F$ contains 2 residual blocks, while network $G$ contains 3. We set the hidden dimension between adjacent residual blocks to 512, and the hidden dimension within each residual block to 128. In the position encoding, we set $L=L_{\mathbf{x}}=12$ and $L_t=4$, while in the sine activation function, $\omega_0=64$.} Both networks are trained using the Adam optimizer with weight decay $10^{-4}$ and a batch size of 4096 voxels. Network $F$ undergoes training for 1200K steps, while network $G$ is trained for 800K steps. The learning rate is initialized to $10^{-3}$ and decreases by a factor of $0.1$ every quarter of the entire training process. During training of $F$, we set $\alpha=\beta=0.5$. We quantize the optimized network parameters of $F$ and $G$ with step sizes $\Delta_F=1/1024$ and $\Delta_G=1/4096$, respectively. Finally, we employ DeepCABAC \cite{wiedemann2020deepcabac} to losslessly compress the quantized parameters.

We adjust parameter settings to achieve different bit rates. In each experiment, we keep the regularization strengths $\lambda_F$ and $\lambda_G$ identical. Therefore, we use a single notation $\lambda$ to denote either one.

\begin{itemize}
\item For i-NeRC$^3$, we adjust $\lambda$ to achieve different bit rates. We set $\lambda$ as $1,5,20,50$.
\item For r-NeRC$^3$, we set $\lambda$ as $5,20,50$.
\item For c-NeRC$^3$, we set $P=3$ and $\lambda=1$, and adjust $T$ for various bit rates. The value of $T$ is selected from $4,8,16,32$.
\item For 4D-NeRC$^3$, we use different $(T,\lambda)$ pairs to attain the corresponding bit rates. The $(T,\lambda)$ pair is selected from $(2,1),(4,1),(8,1),(32,1),(4,10),(16,10)$.
\end{itemize}

\subsubsection{Baselines}

We primarily compare our method with the latest version of MPEG PCC test models, including TMC13-v23.0-rc2 for G-PCC \cite{mpegpcctmc13} and TMC2-v25.0 for V-PCC \cite{mpegpcctmc2}. Since G-PCC employs separate geometry and attribute coding, we utilize G-PCC (octree) and G-PCC (trisoup) for geometry compression, and G-PCC (RAHT) for attribute compression. For V-PCC, we set the mode to \textit{all-intra} for intra-frame compression and \textit{random-access} for inter-frame compression. Parameter settings of these baselines follow the common test conditions \cite{schwarz2018ctc}.

{ Additionally, we compare our method with existing INR-based PCC solutions, including NVFPCC \cite{hu2022nvfpcc} for geometry compression and LVAC \cite{isik2022lvac} for attribute compression. Note that LVAC has been compared with Deep-PCAC \cite{sheng2021deeppcac} in \cite{isik2022lvac}, demonstrating superior performance in the YUV PSNR metric. Furthermore, the original implementation of NVFPCC requires manual threshold fine-tuning; to ensure fairness, we incorporate golden section search into NVFPCC. For standalone geometry compression, we also compare with two state-of-the-art learning-based approaches: PCGCv2 \cite{wang2021pcgcv2} for static coding and Akhtar et al. \cite{akhtar2024ddpcc} for dynamic coding. All methods are evaluated on a platform equipped with AMD EPYC 7402 CPU and NVIDIA GeForce RTX 3090 GPU.}

\subsubsection{Evaluation Metrics}

We quantify geometry distortion using point-to-point error (D1) and point-to-plane error (D2) in peak signal-to-noise ratio (PSNR), and evaluate attribute distortion using Y PSNR and YUV (6:1:1) PSNR. Additionally, we employ PCQM \cite{meynet2020pcqm}, a full-reference quality metric for colored 3D point clouds, to quantify the overall distortion of both geometry and attributes, which better aligns with the human visual system. { For dynamic point cloud sequences, the distortion metric is calculated individually for each frame, and the results across all 32 frames are averaged.}

The compressed bit rate is measured by bits per point (bpp), calculated as the total number of bits divided by the number of points in the original point cloud. For performance comparison, we employ Bjøntegaard delta metrics \cite{bjontegaard2001bdpsnr} to quantify improvements in rate-distortion performance across different methods.

\begin{figure}[t]
\vspace{-0.1cm}
\centering
\includegraphics[width=\columnwidth]{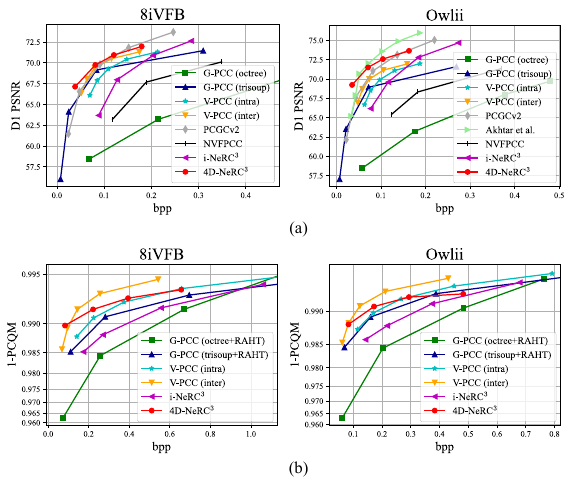}
\vspace{-0.2cm}
\caption{ Rate-distortion comparison of different methods for (a) geometry compression alone, and (b) joint compression of geometry and attributes. The detailed results are available in the supplementary material. }
\label{fig:results}
\vspace{-0.1cm}
\end{figure}

\begin{figure}[t]
\vspace{-0.1cm}
\centering
\includegraphics[width=\columnwidth]{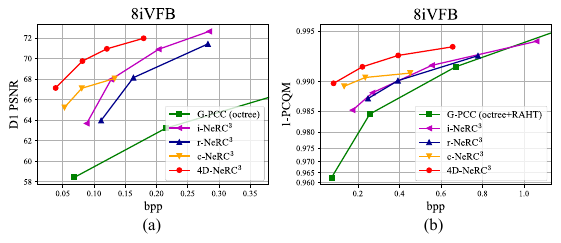}
\vspace{-0.2cm}
\caption{Rate-distortion comparison of the four proposed methods for (a) geometry compression alone, and (b) joint compression of geometry and attributes. The anchor G-PCC (octree) and G-PCC (octree+RAHT) serve as references. }
\label{fig:results_proposed}
\vspace{-0.1cm}
\end{figure}

\subsection{Performance}

\subsubsection{Static PCC}

We compare the performance of i-NeRC$^3$ against the baselines when compressing static point clouds. { The performance comparisons for geometry compression alone are presented in Table \ref{tab:bdpsnr_geom}. We first focus on the results on the 8iVFB and Owlii datasets. As shown, i-NeRC$^3$ outperforms G-PCC (octree) and NVFPCC in both D1 and D2 metrics across all test point clouds. However, i-NeRC$^3$ suffers significant performance losses compared with G-PCC (trisoup) and VPCC (intra). This stems from the inferior performance at low bit rates, as evidenced by the rate-distortion curves in Fig. \ref{fig:results}(a). Nevertheless, at higher bitrates, i-NeRC$^3$ outperforms G-PCC (trisoup), particularly on the Owlii dataset. PCGCv2 achieves the best performance on the 8iVFB and Owlii datasets.

Regarding the four static scenes, the conclusions differ slightly. i-NeRC$^3$ performs poorly on the \textit{Facade} scene due to numerous unclosed holes on the scene surface, which hinder INR overfitting. However, on \textit{Shiva} and \textit{Staue}, i-NeRC$^3$ achieves the best D1 metric performance. In contrast, PCGCv2 exhibits poor D1 metric performance on these two scenes. This likely stems from the rough surfaces of these scenes, which differ significantly from the smooth objects used for training PCGCv2.

In Table \ref{tab:bdpsnr_attr} and \ref{tab:bdpcqm_attr}, we evaluate i-NeRC$^3$ and LVAC on 8iVFB and compare their attribute compression capabilities. To isolate attribute compression, we assume geometry can be losslessly recovered and slightly modify our framework to encode only color information. Clearly, i-NeRC$^3$ outperforms LVAC across all point clouds. LVAC performs poorly on \textit{redandblack} in Y (YUV), as it generates incorrect colors for some points. Both i-NeRC$^3$ and LVAC exhibit inferior performance compared to G-PCC (RAHT).

Table \ref{tab:bdpsnr_joint} and \ref{tab:bdpcqm_joint} demonstrate the overall performance comparison of compressing both geometry and attributes. When evaluated on 8iVFB and Owlii, i-NeRC$^3$ outperforms GPCC (octree+RAHT) in all metrics. Compared with G-PCC (trisoup+RAHT) and V-PCC (intra), i-NeRC$^3$ achieves performance gains in D1 metric but incurs performance losses in Y (YUV) and PCQM metrics. The corresponding rate-distortion curves in PCQM are shown in Fig. \ref{fig:results}(b).
}

\subsubsection{Dynamic PCC}

Given that we have proposed three methods to leverage temporal correlations, namely r-NeRC$^3$, c-NeRC$^3$, and 4D-NeRC$^3$, we first demonstrate the performance comparison between these extended methods and the intra-frame compression method i-NeRC$^3$. All the four methods are tested on the 8iVFB dataset. As shown in Fig. \ref{fig:results_proposed}, r-NeRC$^3$ exhibits poor performance in the D1 metric due to significant difference in voxel occupancies between adjacent point cloud frames. In contrast, c-NeRC$^3$ and 4D-NeRC$^3$ successfully reduce temporal redundancy, outperforming i-NeRC$^3$. Furthermore, 4D-NeRC$^3$ achieves the best performance among the four methods, while c-NeRC$^3$ is only suitable for low bit rates due to the large value of $T$ and still falls short of 4D-NeRC$^3$ in performance.

We now compare 4D-NeRC$^3$, which performs best among the three extended methods, with the baseline methods. { The performance improvement for geometry compression is shown in Table \ref{tab:bdpsnr_geom}. 4D-NeRC$^3$ outperforms all static coding methods and V-PCC (inter). The rate-distortion curves are shown in Fig. \ref{fig:results}(a). Although 4D-NeRC$^3$ exhibits an average performance loss compared to Akhtar et al., it still demonstrates competitive performance at low bit rates.

Regarding attribute compression alone, as shown in Table \ref{tab:bdpsnr_attr} and \ref{tab:bdpcqm_attr}, 4D-NeRC$^3$ outperforms the anchor G-PCC (RAHT) by leveraging inter-frame similarity.

For the overall performance of compressing both geometry and attributes, the comparison results are presented in Table \ref{tab:bdpsnr_joint} and \ref{tab:bdpcqm_joint}. 4D-NeRC$^3$ outperforms G-PCC (octree+RAHT), G-PCC (trisoup+RAHT) and V-PCC (intra) in all metrics, validating its effectiveness in eliminating temporal redundancy. Although 4D-NeRC$^3$ exhibits an average performance loss compared to V-PCC (inter), it demonstrates superior performance on \textit{soldier} and achieves comparable results on \textit{exercise}. The rate-distortion curves in PCQM are shown in Fig. \ref{fig:results}(b).
}

\begin{figure*}[t]
\vspace{-0.1cm}
\centering
\includegraphics[width=2\columnwidth]{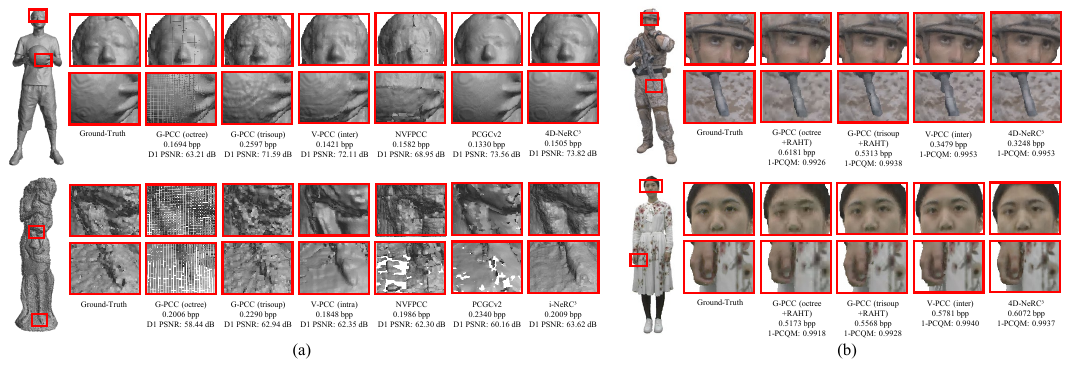}
\vspace{-0.2cm}
\caption{Qualitative visualization. { (a) Reconstruction results of \textit{basketball\_player} and \textit{Staue\_Klimt} for geometry compresion.} (b) Reconstruction results of \textit{soldier} and \textit{model} for joint geometry and attribute compression. }
\label{fig:visual}
\vspace{-0.1cm}
\end{figure*}

\subsubsection{Qualitative Visualization}

For geometry compression, we visualize the point clouds reconstructed by different methods in Fig. \ref{fig:visual}(a). G-PCC (octree) generates sparse point clouds with much fewer points, making it difficult to form continuous surfaces and resulting in a lack of geometry details. Although the reconstructions of G-PCC (trisoup) can approximate water-tight surfaces, they suffer from visible roughness. The point clouds reconstructed by V-PCC exhibit noticeable seams on the face and the basketball in \textit{basketball\_player}. { The reconstructions of NVFPCC similarly produce cube-shaped seams due to its independent processing of each cube. Although PCGCv2 produces satisfactory reconstructions on smooth objects like \textit{basketball\_player}, it performs poorly on rough scenes like \textit{Staue\_Klimt} due to significant differences from PCGCv2's training dataset. In contrast, our implicit representation of point clouds excels in constructing complete and detailed geometry for both types of point clouds.}

For joint compression of both geometry and attributes, the point clouds reconstructed by different methods are visualized in Fig. \ref{fig:visual}(b). The reconstructions of G-PCC (octree+RAHT) exhibits a blurred appearance, as its attributes are based on the geometry reconstructed by G-PCC (octree), which lacks spatial details. G-PCC (trisoup+RAHT) exhibits attribute errors near the gun barrel in \textit{soldier}, which may result from from imprecise geometry reconstruction. V-PCC similarly suffers from noticeable geometry distortion when reconstructing the gun barrel in \textit{soldier} and fingers in \textit{model}. Our method not only generates high-quality geometry close to the ground-truth, but also achieves competitive and visually appealing results when considering attributes.

\begin{table}[t]
\centering
\caption{ Average runtime (seconds per frame) of different INR-based methods on 8iVFB. }
\label{tab:runtime}
\begin{threeparttable}
\begin{tabular}{cc|ccccc}
\toprule
\multicolumn{2}{c|}{\multirow{2}{*}{Scenario}} & \multirow{2}{*}{NVFPCC} & \multirow{2}{*}{LVAC} & \multicolumn{2}{c}{i-NeRC$^3$} & 4D-NeRC$^3$ \\
& & & & Full & $1/4\times$ & Full \\
\midrule
\multirow{2}{*}{Geometry} & Enc. & 2924 & - & 9722 & 2588 & 2935 \\
& Dec. & 11.41 & - & 6.47 & 6.42 & 7.63 \\
\midrule
\multirow{2}{*}{Attributes} & Enc. & - & 4816 & 6726 & 1633 & 1677 \\
& Dec. & - & 13.75 & 0.21 & 0.22 & 0.27 \\
\bottomrule
\end{tabular}
\begin{tablenotes}
\footnotesize
\item ``Full" means training network $F$/$G$ for 1200K/800K steps.
\item ``$1/4\times$" means training network $F$/$G$ for 300K/200K steps.
\end{tablenotes}
\end{threeparttable}
\end{table}

\begin{figure*}[t]
\vspace{-0.1cm}
\centering
\includegraphics[width=2\columnwidth]{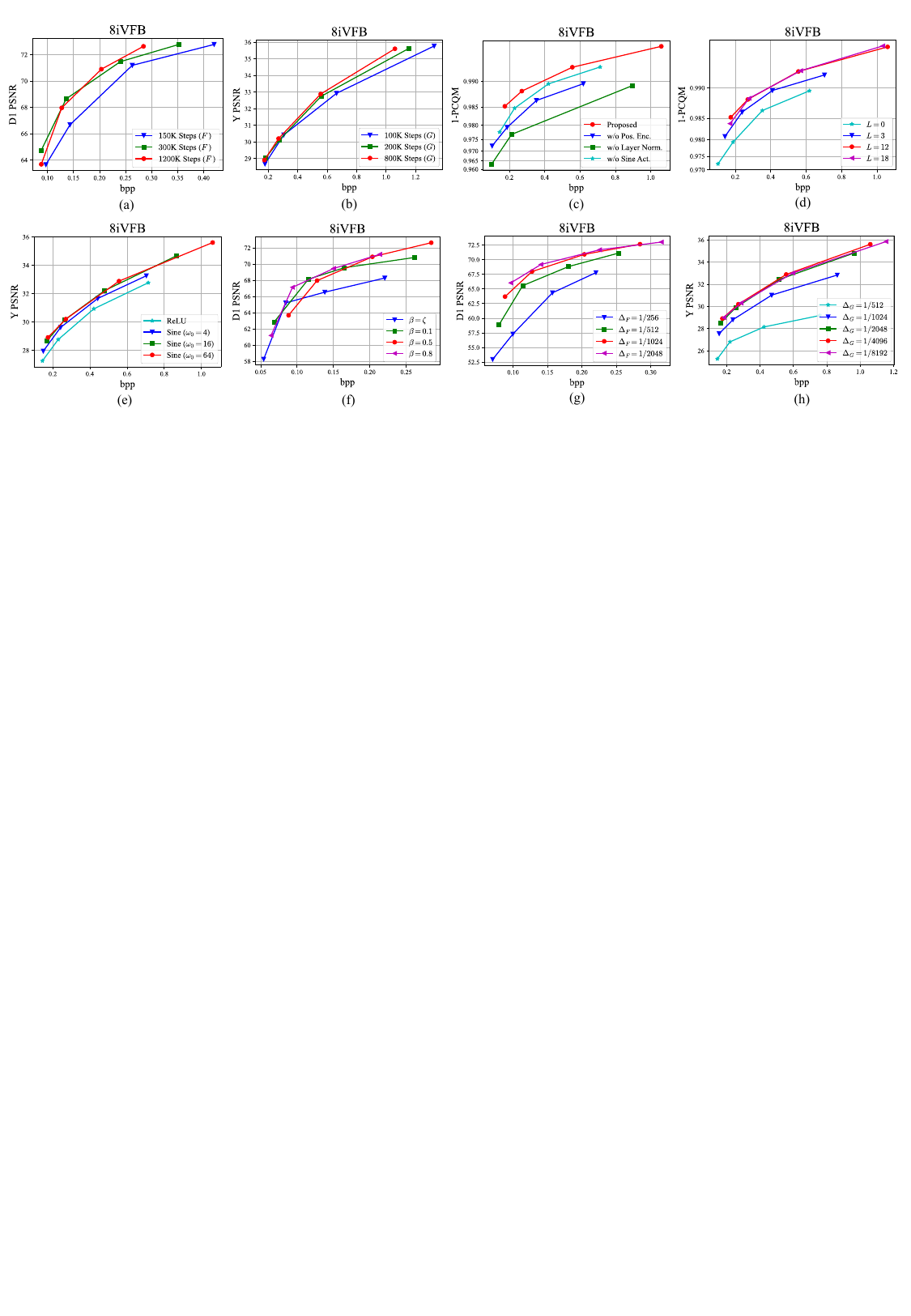}
\vspace{-0.2cm}
\caption{{ (a)-(b) Trade-off between training time and rate-distortion performance. (a) Geometry compression with different numbers of training steps for $F$. (b) Joint compression with different numbers of training steps for $G$.} (c)-(h) Ablation studies. (c) Joint compression using three variants as well as the proposed structure. { (d) Joint compression with different $L$ values. (e) Joint compression with different $\omega_0$ values.} (f) Geometry compression with different ratios of $\beta$. { (g) Geometry compression with different step sizes $\Delta_F$ for network $F$. (h) Joint compression with different step sizes $\Delta_G$ for network $G$.}}
\label{fig:ablation}
\vspace{-0.1cm}
\end{figure*}

{
\subsection{Time Complexity}

As an INR-based method, NeRC$^3$ optimizes neural networks for each point cloud instance individually, resulting in significantly longer encoding runtime compared to non-INR approaches. A detailed complexity analysis is given in the supplementary material. Apparently, encoding time is determined by the number of training steps. In most experiments, we employ a sufficiently large number of training steps to ensure full network convergence, thereby achieving optimal rate-distortion performance. While reducing training steps can shorten runtime, it may compromise performance. To illustrate the trade-off between training time and performance, Fig. \ref{fig:ablation}(a)-(b) displays the rate-distortion performance of i-NeRC$^3$ achieved by training the networks with varying numbers of steps. The performance is evaluated on the 8iVFB dataset. We observe that i-NeRC$^3$ achieves comparable performance after training networks $F$/$G$ for only 300K/200K steps, equivalent to 1/4 of the original 1200K/800K steps, without significant performance degradation.

To ensure a fair comparison, we contrast our approach with INR-based NVFPCC \cite{hu2022nvfpcc} and LVAC \cite{isik2022lvac}. Table \ref{tab:runtime} presents the average runtime of different INR-based methods on the 8iVFB dataset, averaged across multiple bit rate points. 4D-NeRC$^3$ reduces average encoding time compared to i-NeRC$^3$ by optimizing a single network for a group of $T$ frames. Furthermore, by introducing a slight trade-off between training time and performance, our method achieves faster encoding speeds than other INR-based approaches.
}

\subsection{Ablation Studies}

In this subsection, we investigate the effectiveness of different aspects in our proposed INR-based framework i-NeRC$^3$. { We conduct our studies on the 8iVFB dataset.}

\subsubsection{Network Structure}

To demonstrate the benefits of positional encoding, layer normalization, and the sine activation in the network structure, we remove these components separately to obtain three variants of the network structure and compare them with the proposed structure. The rate-distortion curves in Fig. \ref{fig:ablation}(c) evidence that all these components enhance the networks' capacity to achieve higher performance.

{
Next, we further investigate certain hyperparameters related to the network structure. To explore the impact of different $L$ values in the positional encoding, we plot the results in Fig. \ref{fig:ablation}(d). Here, $L=0$ indicates no position encoding is applied. Clearly, larger $L$ values yield superior results compared to smaller ones. However, beyond $L=12$, further increases in $L$ yield negligible performance gains.

Fig. \ref{fig:ablation}(e) shows the results for different $\omega_0$ values in the sine activation, compared against the traditional ReLU activation. Although the input positional encoding has introduced high-frequency variations, replacing ReLU with the sine activation further improves reconstruction quality. However, as $\omega_0$ increases, the bit rate grows in tandem with reconstruction quality, resulting in limited improvement of overall rate-distortion performance. Furthermore, our experiments indicate that when $\omega_0$ becomes excessively large, the network struggles to converge.
}

\subsubsection{Sampling Strategy}

As mentioned before, we introduce a hyperparameter $\beta$ controlling the ratio of occupied voxels in training samples. To validate the proposed sampling strategy and investigate the impact of different $\beta$ values, we perform geometry compression with various $\beta$ values. Fig. \ref{fig:ablation}(f) plots the rate-distortion curves, where $\beta=\zeta$ represents uniform sampling within the volumetric space. We adjusted the $\lambda$ parameter settings to maintain similar ranges of bit rate variation across different $\beta$ conditions. { As shown, sampling with larger $\beta$ yield superior results compared to uniform sampling, particularly at high bit rates, validating the effectiveness of the proposed sampling strategy. However, when $\beta$ varies between 0.1 and 0.8, the rate-distortion performance trend becomes less pronounced. In our experiments, we select $\beta=0.5$ to ensure the method achieves a higher upper bound for reconstruction quality.}

{
\subsubsection{Quantization}

To demonstrate how the step sizes $\Delta_F,\Delta_G$ affect rate-distortion performance, we quantize network parameters using different step sizes and evaluate the rate-distortion behavior. As shown in Fig. \ref{fig:ablation}(g)-(h), larger step sizes (e.g., $1/256$, $1/512$) cause severe parameter distortion, leading to degraded reconstruction quality. As the step size decreases, rate-distortion performance initially improves but eventually reaches a performance ceiling, because the increase in bit rate offsets the gains in reconstruction quality. We select step sizes that enable our method to perform excellently within the target bit rate range.}

%% file: SecVI.tex
In this paper, we have proposed NeRC$^3$, a novel framework for PCC based on INRs. Our approach initially targeted compression of a single point cloud by employing two neural networks to implicitly represent its geometry and attributes, respectively. This representation was followed by parameter quantization and encoding, with the quantized parameters and auxiliary information enabling the reconstruction of a lossy version of the original point cloud. Furthermore, we have extended our method to dynamic PCC, proposing several strategies to exploit temporal redundancy in point cloud sequences, including a 4D spatio-temporal representation approach (4D-NeRC$^3$). Extensive experimental results have validated the effectiveness of our framework.

This work paves a new path for INR-based PCC, supporting the compression of both geometry and attributes for static and dynamic point clouds. Future research may focus on enhancing the rate-distortion performance by exploring more efficient network architectures, such as learnable activation functions \cite{liu2024kan}. { Furthermore, the time and space complexity of INR-based methods can be further reduced. Sparse training \cite{yuan2021mest} demonstrates significant potential, as it has been validated to substantially decrease the training and inference complexity of sparse networks without compromising performance. Additionally, more efficient pre-processing schemes of point cloud geometry could be explored.}